%% file: main.tex
\documentclass{article}
\usepackage[T1]{fontenc}
\usepackage{iclr2027_conference,times}

\usepackage{amsmath,amssymb}
\usepackage{graphicx}
\usepackage{capt-of}
\usepackage{array}
\usepackage{booktabs}
\usepackage{xcolor}
\usepackage{colortbl}
\usepackage{enumitem}
\usepackage{microtype}
\usepackage{url}
\usepackage{hyperref}
\usepackage{fontawesome5}
\usepackage{placeins}
\usepackage{wrapfig}
\newcommand{\samethanks}{\footnotemark[\value{footnote}]}
\newcommand{\sharedaffiliation}{\end{tabular}\par\begin{tabular}[t]{l}}

\definecolor{tabband}{gray}{0.94}

\newcommand{\appref}[1]{\hyperref[#1]{Appendix~\ref*{#1}}}
\newcommand{\secref}[1]{\hyperref[#1]{Section~\ref*{#1}}}
\newcommand{\figref}[1]{\hyperref[#1]{Figure~\ref*{#1}}}
\newcommand{\tabref}[1]{\hyperref[#1]{Table~\ref*{#1}}}
\newcommand{\panelref}[2]{\hyperref[#1]{Figure~\ref*{#1}#2}}

\newenvironment{tablock}
  {\par\addvspace{14pt}\noindent\begin{minipage}{\linewidth}}
  {\end{minipage}\par\addvspace{6pt}}

\newif\ifdraftnotes
\draftnotestrue
\ifdraftnotes

\else

\fi

\setlist[itemize]{nosep,leftmargin=1.4em}

\title{Vision Transformers Learn Word Meaning from Images Alone}

\author{%
Anders Gj{\o}lbye\thanks{Equal contribution.} \And
Sebastian Ray Mason\samethanks \And
Stella Frank \And
Lars Kai Hansen
\sharedaffiliation
Department of Applied Mathematics and Computer Science\\
Technical University of Denmark\\
\texttt{agjma@dtu.dk}}

\iclrfinalcopy

\begin{document}
\maketitle
\lhead{Preprint}
\linespread{0.975}\selectfont 

\input{sections/0_abstract}
\input{sections/1_introduction}
\input{sections/2_related_work}

\input{sections/3_methods}

\input{sections/4_lexical_structure}
\input{sections/5_word_concept_alignment}
\input{sections/6_finetuning_and_functional_use}
\input{sections/7_discussion}
\input{sections/8_conclusion}

\section*{Acknowledgments}
This work was supported by the Novo Nordisk Foundation grant NNF22OC0076907 ``Cognitive spaces -- Next generation explainability'' and the Pioneer Centre for AI, DNRF grant number P1. It was also supported by the Danish Data Science Academy, which is funded by the Novo Nordisk Foundation (NNF21SA0069429) and VILLUM FONDEN (40516).

\bibliographystyle{iclr2027_conference}
\bibliography{refs}

\appendix
\clearpage\input{appendices/A_data_and_rendering}
\clearpage\input{appendices/B_measures}
\clearpage\input{appendices/C_lexical_space_analyses}
\clearpage\input{appendices/D_word_concept_analyses}
\clearpage\input{appendices/E_finetuning_and_reproducibility}

\end{document}

%% file: sections/0_abstract.tex
\begin{abstract}
Can a vision model learn what words mean from pixels alone? We show that the two largest DINOv3 vision transformers organize words rendered in images by word meaning, despite being pretrained using self-supervision on images only, without paired captions, class labels, or other language supervision. We derive word representations from embedding differences between word images and matched controls, then test whether semantically related words are neighbors within each model's embedding space, while controlling for spelling similarity. We find that brand names in the same industry group together, as do nouns from the same semantic domain. Country-capital pairs exhibit consistent vector offsets, similarly to word embeddings trained on text. These word representations also align with the representations of the visual object categories they name. DINOv3-H and DINOv3-7B are, to our knowledge, the first self-supervised vision models to show substantial semantic structure for rendered words. Comparisons with text and speech encoders reveal shared structure among representations of the same words. Contrastive alignment with either encoder further strengthened DINOv3-H's word-level semantic structure, while leaving the other finetuned vision models largely unchanged. Image-only training can therefore teach vision models not just what words look like, but also how their meanings relate.
\end{abstract}

%% file: sections/1_introduction.tex
\section{Introduction}

Can knowledge of semantic relations among words emerge in models trained on images alone? Self-supervised vision models learn representations organized by the semantic categories of depicted objects \citep{oquab2024dinov2,simeoni2025dinov3}. Written words can also appear within their training images, but are encountered only as pixels, without paired captions, class labels, or other language supervision \citep{fan2025scaling}. Distinguishing these words by appearance does not establish knowledge of how their meanings relate. Word embeddings trained on text can encode such relations in their geometry \citep{mikolov2013distributed}, based on co-occurrence patterns in the textual input.
We ask whether image-only training can produce similar structure.
For example, is a vision model's representation of country names systematically offset with those of the capitals?

Training without language supervision does not mean training without exposure to language: written words may still appear as pixels within images. CLIP, by contrast, learns to recognize text through image-caption training \citep{radford2021clip}, and in that model certain neurons respond to the same concepts either depicted or written \citep{goh2021multimodal}. However, earlier comparisons found that DINOv2's predictions were unaffected by the meaning of overlaid words \citep{wang2025textural}, and that its representations grouped images by visual content rather than written word semantics \citep{jalali2025explainable}.
More recently, \citet{mason2026mentalrotation} showed that DINOv3 representations distinguish rotated from mirrored text. While this result concerns visual structure rather than word meaning, it suggests that DINOv3 represents written strings more richly than earlier results might imply. We therefore ask whether its representations also capture semantic relations among words.

Self-supervised vision models already learn semantic categories of depicted objects \citep{oquab2024dinov2,simeoni2025dinov3}. But recognizing an object, or distinguishing one written word from another, does not establish knowledge of semantic relations among words. We find that this capability emerges in the largest DINOv3 models but is largely absent from the earlier self-supervised vision models we test. Thus, image-only training can recover some of the semantic structure typically studied through language, despite having no objective that requires the model to read or match language. More broadly, models can acquire capabilities that their training objective never explicitly targets when the relevant structure is present in the training data. Understanding such unexpected capabilities is therefore a general problem in large-scale pretraining, not one specific to text recognition.

\begin{figure}[t]
\centering
\begin{minipage}[t]{\dimexpr\textwidth-206pt\relax}
\vspace{0pt}
\scriptsize
\setlength{\defaultaddspace}{2pt}
\renewcommand{\arraystretch}{1.0}
\setlength{\aboverulesep}{0.3ex}
\setlength{\belowrulesep}{0.3ex}
\input{tables/figure1a_neighbours}
\end{minipage}\hfill
\begin{minipage}[t]{199.98bp}
\vspace{0pt}
\includegraphics{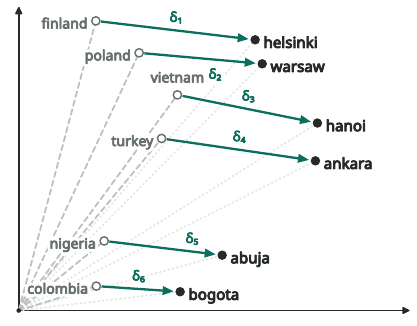}%
\setlength{\unitlength}{1bp}%
\end{minipage}
\caption{\textbf{A vision transformer trained on only images ranks the capital first.} \textbf{(a)} Three nearest rendered words per query, with cosine similarities between centered word directions shown at left. Bold marks meaning partners or same-industry brands.
\textbf{(b)}~Country-to-capital offsets in DINOv3-H, illustrating the vector-offset analogy relations studied by \citet{mikolov2013distributed}. Gray lines indicate centered word directions $\tau$; their common origin is for layout purposes only. The six pairs, hand-picked from 60, illustrate the relation rather than quantify its consistency. See \appref{app:projection}.
}
\label{fig:nearest}
\end{figure}

 We analyze rendered words using word directions: embedding differences between word images and matched controls, averaged across backgrounds and centered over the vocabulary. Controlling for spelling similarity and rendering conditions, we find: (1)~In DINOv3-H and DINOv3-7B, country-capital pairs are neighbors with consistent offsets, brands and nouns group by category, and synonyms are retrieved above chance (\figref{fig:nearest}). These models perform comparably to CLIP-B on several measures, while earlier self-supervised models primarily group words by visual similarity. To our knowledge, no prior work has shown that vision models pretrained without language supervision encode semantic relations among rendered words. (2)~In these large DINOv3 models, word directions align with centered mean image embeddings of the corresponding visual object, without caption supervision or a fitted mapping. (3)~Comparisons with text and speech encoders reveal shared representational structure. Contrastive finetuning on image-text or image-speech pairs, using images with no detected text, improves synonym retrieval and noun-category grouping in DINOv3-H, but changes the other finetuned models minimally.
 
%% file: tables/figure1a_neighbours.tex
\definecolor{tabband}{gray}{0.94}
\begingroup
\def\tablebody{%
\toprule
 & Capital & Synonym & Category \\
\cmidrule(l{2pt}r{2pt}){2-2}\cmidrule(l{2pt}r{2pt}){3-3}\cmidrule(l{2pt}r{2pt}){4-4}
Model & germany & petrol & porsche \\
\midrule
\rowcolor{tabband}DINOv3-H & .87 \textbf{berlin} & .68 \textbf{gasoline} & .72 \textbf{ferrari} \\
\rowcolor{tabband} & .62 vienna & .46 patrol & .26 terrace \\
\rowcolor{tabband} & .57 edelweiss & .29 patio & .23 germany \\
\rowcolor{tabband}DINOv3-7B & .78 \textbf{berlin} & .70 patrol & .45 \textbf{ferrari} \\
\rowcolor{tabband} & .58 austria & .68 \textbf{gasoline} & .41 \textbf{volkswagen} \\
\rowcolor{tabband} & .57 vienna & .40 engine & .29 \textbf{renault} \\
\addlinespace
Web-DINO-1B & .70 hungary & .61 privet & .52 polished \\
 & .67 guyana & .52 polite & .48 carlsberg \\
 & .65 generous & .51 gerbil & .47 precise \\
\addlinespace
\rowcolor{tabband}DINOv2-G & .62 meatloaf & .55 patrol & .54 tadpole \\
\rowcolor{tabband} & .62 gatorade & .54 barrel & .53 careless \\
\rowcolor{tabband} & .61 heineken & .52 trench & .51 colombia \\
\addlinespace
CLIP-B & .62 \textbf{berlin} & .44 patrol & .54 porch \\
 & .44 austria & .36 pepsi & .46 perch \\
 & .43 poland & .32 chevrolet & .39 epoch \\
\bottomrule
}%
\setlength{\tabcolsep}{0pt}%
\sbox0{\begin{tabular}{llll}\tablebody\end{tabular}}%
\ifdim\wd0<\linewidth%
\setlength{\tabcolsep}{\dimexpr(\linewidth-\wd0)/\numexpr2*4\relax\relax}%
\else\setlength{\tabcolsep}{0pt}\fi%
\begin{tabular}{llll}\tablebody\end{tabular}
\endgroup

%% file: sections/2_related_work.tex
\section{Related Work}

\paragraph{Self-supervised vision and language supervision.} Self-distillation produces representations organized by semantic category \citep{caron2021dino,oquab2024dinov2,simeoni2025dinov3}, while masked reconstruction \citep{he2022mae} places greater emphasis on local visual features and texture \citep{park2023ssl}. At scale, self-supervised image encoders can match the performance of language-supervised encoders when used in visual question answering systems \citep{fan2025scaling}. A controlled comparison at a smaller scale, holding architecture and training images fixed, finds that caption supervision strengthens representations of fine-grained object categories and of text \citep{data2025supervision}.

\paragraph{Rendered words in vision models.} CLIP learns to recognize text in images as a by-product of image-text contrastive training \citep{radford2021clip}. Certain neurons respond to a concept whether it is depicted or written \citep{goh2021multimodal}. Learned projections can disentangle written and visual concepts in CLIP, and its image encoder matches rendered class names to corresponding natural images \citep{materzynska2022disentangling}. Representations of rendered text in CLIP's image encoder show semantic alignment with a language model comparable to that of CLIP's text encoder \citep{vennam2024semantics}. CLIPPO uses a single image encoder for images and rendered alt-text, but still trains contrastively on image--alt-text pairs \citep{tschannen2023clippoimageandlanguageunderstandingpixels}. Earlier comparisons find that DINOv2's predictions are unaffected by the meaning of overlaid words \citep{wang2025textural}, and that its embeddings group images by visual content rather than the written word \citep{jalali2025explainable}. Our aim is to characterize semantic relations learned without language supervision, not to propose a method for language understanding.

\paragraph{Representational convergence.} The Platonic Representation Hypothesis proposes that models trained on different modalities converge toward shared representations as they scale \citep{huh2024platonic}. However, image-caption neighborhood agreement weakens on larger evaluation sets and mainly reflects coarse semantic overlap \citep{koepke2026plato}. After correcting for model-size biases, \citet{groger2026aristotelian} find significant agreement in local neighborhoods but little evidence of global convergence under spectral measures. %

\paragraph{The geometry of word meaning.} Words used in similar contexts tend to have related meanings \citep{harris1954distributional}; distributional semantic models leverage contextual co-occurrence to learn a representation of word meanings. In such models, word embeddings encode semantic relations as similar vector offsets, for example between countries and their capitals \citep{mikolov2013distributed}. We ask whether comparable geometric structure emerges in representations of rendered words learned from images alone. Because word form can weakly covary with meaning, as documented for sound \citep{monaghan2014arbitrary}, we test these relations while controlling for spelling similarity. Separately, we test visual word-referent grounded alignment~\citep{bender2020climbing}, by testing whether word representations align with representations of the depicted object categories they name.

%% file: sections/3_methods.tex
\section{Methods}
\label{sec:methods}

Our experiments first test whether vision models represent semantic relations among rendered words, controlling for spelling and rendering effects (\secref{sec:lexical}). We then compare word directions with representations of the visual object categories they name (\secref{sec:alignment}). Finally, we compare the semantic structure of word directions with representations from text and speech encoders, and test whether contrastive finetuning toward these representations through written or spoken captions strengthens that structure (\secref{sec:encoders}).\footnote{Code is available at \href{https://github.com/gjoelbye/vision-word-semantics}{\faGithub\,gjoelbye/vision-word-semantics}.}

\subsection{Models}
\label{sec:models}
We measure fourteen models (\tabref{tab:models}): four DINOv3 models and Web-DINO-1B, trained without labels or captions on images alone, DINOv3-7B by self-distillation and the smaller three by distillation from it \citep{simeoni2025dinov3,fan2025scaling}; three DINOv2 models, the predecessors to DINOv3 \citep{oquab2024dinov2}; three MAE models, trained by masked reconstruction \citep{he2022mae}; CLIP-B and SigLIP2-B, trained with language supervision on image-caption pairs \citep{radford2021clip,tschannen2025siglip2}; and an untrained ViT-B as the baseline for visual form alone. The encoder $f$ maps an image $x$ to its final-layer class token, or for SigLIP2-B to the mean of its patch tokens, with no projection head and no per-image normalization.

\subsection{Words and images}
\label{sec:rendering}

\begin{figure}[t]
\centering
\includegraphics[width=\textwidth]{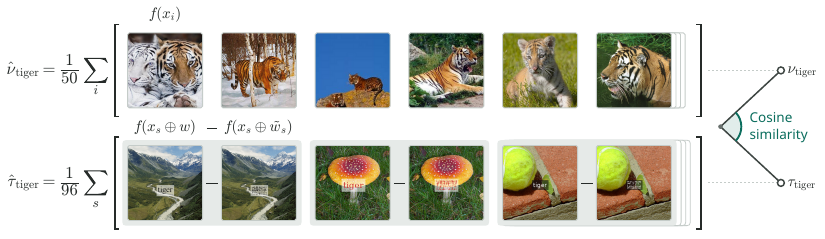}
\caption{\textbf{Visual and word directions for \emph{tiger}.} The upper row illustrates the uncentered visual class mean $\hat{\nu}_{\mathrm{tiger}}$ of Eq.~\eqref{eq:nu}, corresponding to the mean final-layer representation over 50 ImageNet-1k validation images for a class. 
The lower row is the uncentered word difference $\hat{\tau}_{\mathrm{tiger}}$ of Eq.~\eqref{eq:tau}, averaged over 96 backgrounds. The word image $x_s \oplus w$ shows word $w$ on background image $x_s$; its control $x_s \oplus \tilde{w}_s$ has the same background, with the letters' opacity mask replaced by a phase-scrambled patch. Stacked card edges mark terms not drawn. Centering over all 1,000 classes gives the visual direction $\nu_{\mathrm{tiger}}$; centering over the 200 class names gives the word direction $\tau_{\mathrm{tiger}}$. The rays show the cosine similarity between these directions, with schematic lengths and angle.}
\label{fig:datasets}
\end{figure}

Two vocabularies are used. The \emph{lexical vocabulary} tests whether related words are neighbors (\secref{sec:lexical}) and supports comparisons with text and speech encoders (\secref{sec:encoders}). The \emph{class-name vocabulary} tests whether a rendered class name is close to images of the class it names (\secref{sec:alignment}). \tabref{tab:wordlists} in \appref{app:data} gives every list, its size, and an example.

\paragraph{Lexical vocabulary.}
The lexical vocabulary contains 1,481 words and names and 80 pseudowords, all lowercase and three to ten letters long. The words come mainly from WordNet, with a target frequency of at least one occurrence per million words to favor familiar words. The vocabulary is divided into eight lists, used in three ways:

\emph{Relation lists.}
Five lists are generated for partner retrieval: whether a word's partner is among its ten nearest neighbors (\secref{sec:readouts}). Three lists use triplets containing an anchor, a synonym, and a competitor. In the one-edit list (1), the competitor differs from the anchor by one letter insertion, deletion, or substitution (150 items; anchor: \emph{torch}, synonym: \emph{flashlight}, competitor: \emph{touch}). This creates a controlled competition between semantic relatedness and close orthographic form. In the antonym list (2), the competitor is an antonym (50 items; \emph{brave, bold, cowardly}). In the negation list (3), the competitor is a similarly spelled word that typically corresponds to the anchor's unnegated base (35 items; disappear, vanish, appear); exceptions are described in \appref{app:data}. The singular--plural list (4) contains 40 pairs related in both form and meaning (\emph{road--roads}). The country--capital list (5) contains 60 pairs expressing a factual relation (\emph{germany--berlin}). Retrieval always uses all 1,560 other vocabulary items as candidates, not only items from the same list.

\emph{Noun and brand lists.}
The noun list (6) contains 384 common nouns from six categories (animal, device, food, person, plant, structure). The brand list (7) contains 192 brand names from six industries. These lists are used for neighbor purity: the fraction of a word's ten nearest neighbors that belong to the same category or industry (\secref{sec:readouts}).

\emph{Pseudowords.}
The 80 pseudowords (8) are constructed nonwords and are never used as queries. They serve as additional retrieval candidates and are included in the mean used for centering.
\paragraph{Class-name vocabulary.}
The class-name vocabulary contains 200 single-word ImageNet-1k class names, each chosen to clearly name one class. We prefer classes whose images contain little detected text, reducing the chance that a rendered name matches writing already present in the class images. The names are rendered on the same 96 backgrounds. For each of the 1,000 ImageNet classes, a visual direction is computed from the 49,904 validation images not used as backgrounds. Each name's word direction is then compared with the visual direction of the class it names (\appref{app:concepts}). Forty-three names also occur in the lexical vocabulary, but their word directions differ because the two vocabularies use different centering means (\secref{sec:word_vectors}).

\paragraph{Form controls.} The spelling controls test whether shared letters explain similarity between related words. We screen meaning pairs for inflectional and derivational relatives. Their average letter overlap within each relation is close to that of random pairs matched in length.
We group words with similar character subsequences into four \emph{spelling clusters}, requiring equal numbers from each category or industry in every cluster. Cluster membership alone therefore cannot predict those labels. We also measure category membership among neighbors within each cluster and compare the scores with category labels shuffled among words of similar length (\appref{app:category}).

\paragraph{Rendering.} Every word, name, and pseudoword is rendered on the same 96 ImageNet-1k validation backgrounds with no detected text. The 24 settings combine six fonts with four vertical position bands and vary size, color, and backing. Each setting is used on four backgrounds. Typography and text center are fixed across words on each background, so words are compared under the same rendering conditions.

\paragraph{Control image.} Each word image has a control with the same background, position, color, and backing. Fourier phase randomization replaces the letters' opacity mask with a textured patch in the same bounding box. Subtracting their representations measures the response to intact lettering relative to this matched overlay. Averaging these differences across backgrounds and subtracting the vocabulary mean gives one word direction (\appref{app:data}).

\subsection{Word and visual directions}
\label{sec:word_vectors}
\label{sec:word_class}

\paragraph{Word directions.} Let $S$ index the 96 backgrounds and $x_s$ denote background image $s$. For a word, name, or pseudoword $w$, write $x_s \oplus w$ for its word image, where $\oplus$ denotes rendering with the settings assigned to that background. The matched control is $x_s \oplus \tilde{w}_s$, where $\tilde{w}_s$ denotes the phase-scrambled opacity mask of the lettering for word $w$ on background $s$. The \emph{word direction} $\tau_w$ is the average difference between the word and control representations, centered over vocabulary $V$:
\begin{equation}
\label{eq:tau}
\hat{\tau}_w \;=\; \frac{1}{|S|}\sum_{s \in S} \bigl[\, f(x_s \oplus w) - f(x_s \oplus \tilde{w}_s) \,\bigr], \qquad \tau_w \;=\; \hat{\tau}_w - \frac{1}{|V|}\sum_{v \in V} \hat{\tau}_v,
\end{equation}
Here, $\hat{\tau}_w$ is the intermediate, uncentered difference vector. For lexical and cross-encoder comparisons, $V$ contains 1,481 words and names and 80 pseudowords. For word-class comparisons, it contains the 200 class names. The uncentered estimate is a mean of paired differences, as in difference-in-means probes \citep{rimsky2024caa,marks2024geometry} and concept directions under the linear representation hypothesis \citep{park2024lrh}. The matched control is used to improve signal-to-noise by controlling for background and rendering effects. Directly averaging the word-image embeddings without control subtraction leads to the same qualitative conclusions.

\paragraph{Visual directions.} For a class $c$, let $I_c$ index its retained class images $x_i$. Their uncentered class mean is $\hat{\nu}_c$. Centering it over the set $C$ of all 1,000 classes gives the \emph{visual direction} $\nu_c$:
\begin{equation}
\label{eq:nu}
\hat{\nu}_c \;=\; \frac{1}{|I_c|}\sum_{i \in I_c} f(x_i), \qquad \nu_c \;=\; \hat{\nu}_c - \frac{1}{|C|}\sum_{c' \in C} \hat{\nu}_{c'}.
\end{equation}
Centering over all 1,000 classes keeps the reference fixed when we choose the 200 classes with single-word names. The resulting visual direction is the difference between that class's mean and the mean of all the classes \citep{kim2018tcav}. \figref{fig:datasets} compares the word and visual directions for \emph{tiger}.

\subsection{Measures}
\label{sec:readouts}

We use cosine similarity to find nearest neighbors. For the noun and brand lists, \emph{neighbor purity} is the average fraction of a word's ten nearest neighbors that belong to the same category or industry. For noun purity, neighbors are drawn from the other 383 nouns; for brand purity, from the other 191 brands. For the five relation lists, \emph{partner retrieval} asks whether a word's designated partner, such as its synonym or capital, is among its ten nearest neighbors in the full lexical vocabulary. The query itself is excluded. \emph{Top-10 accuracy} is the fraction of queries whose designated target is among the ten nearest candidates.

For word--class alignment, \emph{matched cosine similarity} is the average similarity between each class name's word direction and the visual direction of the class it names. Word-to-class retrieval ranks all 1,000 visual directions for each of the 200 class names, while class-to-word retrieval ranks the 200 word directions for each named class. Both use top-10 accuracy. All comparisons are made directly in the same embedding space; no mapping between word and visual directions is fitted. See \appref{app:measures} for further details.

\subsection{Alignment and finetuning}
\label{sec:cross_encoder}

\paragraph{Encoders.} We compare vision models' word directions with word representations from Qwen3-Embedding-0.6B, -4B, and -8B for text \citep{zhang2025qwen3embedding}, and w2v-BERT~2.0 for speech \citep{seamlessm4t2023}. For text, each word, name, and pseudoword is placed in the same 96 neutral contexts, such as ``The word is \emph{w}.'' We average the resulting text embeddings. For speech, four shared carrier sentences are spoken in eight voices at three rates, giving 96 utterances per word. We pool only the frames in which the word is spoken, then average over utterances. Both averages are centered over the full vocabulary of 1,561 words, names, and pseudowords. \appref{app:encoders} gives the templates, voices, and pooling details.

\paragraph{Measure of alignment.} We test whether vision, text, and speech encoders agree about which words are related, despite having different input modalities. For each encoder, we form a matrix of cosine similarities between its centered vectors for the same vocabulary. Mutual 10-NN alignment measures the average fraction of each word's ten nearest neighbors shared by two encoders \citep{huh2024platonic}. Linear centered kernel alignment (linear CKA) compares the full pattern of pairwise similarities, using the normalized inner product of the two centered matrices \citep{kornblith2019cka}. \appref{app:measures} defines both scores, and \appref{app:alignment_tests} explains how agreement is compared with chance.

\paragraph{Finetuning.} To test whether language supervision can build on existing word relationships, we finetune self-supervised vision models to match COCO images to either their written captions \mbox{\citep{lin2014coco,chen2015cococaptions}} or synthesized readings from SPEECH-COCO \citep{havard2017speechcoco}. We exclude images with detected text to limit direct supervision from writing in images. The text and speech runs use the same images and captions. A learned linear projection maps the vision model's image class token to the embedding dimension of the frozen text encoder (Qwen3-Embedding-0.6B) or speech encoder (w2v-BERT~2.0). Symmetric InfoNCE loss \citep{radford2021clip} trains the vision model and projection to match the corresponding caption's text or speech vector. We discard the projection and repeat the same word tests after finetuning, so only the vision model's weights have changed (\appref{app:finetune_recipe}).

%% file: sections/4_lexical_structure.tex
\begin{table}[t]
\centering
\footnotesize
\setlength{\defaultaddspace}{2pt}
\renewcommand{\arraystretch}{0.85}
\setlength{\aboverulesep}{0.3ex}
\setlength{\belowrulesep}{0.3ex}
\setlength{\abovecaptionskip}{5pt}
\setlength{\belowcaptionskip}{\abovecaptionskip}
\setlength{\abovecaptionskip}{0pt}
\caption{\textbf{DINOv3-H and DINOv3-7B are comparable to CLIP-B.}
Neighbor purity is the fraction of the ten nearest neighbors sharing a noun category or brand industry; top-10 accuracy is the fraction of paired targets retrieved among the ten nearest vocabulary items.
Underlining marks a value above the untrained ViT-B by more than twice the standard error of the difference, bold marks column maxima, and small numbers are standard errors over words or pairs (\appref{app:statistics}).}
\label{tab:models}
\input{tables/table1_models}
\end{table}

\section{Lexical Structure}
\label{sec:lexical}

Our first experiment evaluates whether word directions of rendered words are semantically structured; results are shown in \tabref{tab:models}, \panelref{fig:nearest}{a}, and \appref{app:lexical}.

\paragraph{Semantic relations.} DINOv3-H and DINOv3-7B group rendered words by meaning. In both, 60\% of countries have their capital among their ten nearest neighbors (c.f.~55\% in CLIP-B, with language supervision, but 0\% for DINOv2), and nouns and brands lie near words of their own category about as often as in CLIP-B. DINOv3-L shows a weaker version of the same structure. The other eight self-supervised models, Web-DINO-1B among them, perform on par with the untrained ViT-B, using only structural inductive bias.
However, synonym retrival is limited in vision models: top-10 accuracy is 20\% in CLIP-B and 10\% in DINOv3-7B, compared with 69--74\% in the Qwen3 text encoders given the vocabulary as text. Across depth, noun purity rises from mid-depth in DINOv3-H, DINOv3-7B, CLIP-B, and SigLIP2-B, while capital retrieval emerges only in the last two measured depths of DINOv3-H, DINOv3-7B, and CLIP-B; the remaining models stay near the untrained ViT-B baseline, except DINOv3-L at the final depth (\appref{app:lexical_depth}).

\paragraph{Spelling.}
Sensitivity to visual overlap between words does not by itself imply a representation of spelling. Every model retrieves one-edit words, and even the untrained ViT-B achieves 33\% top-10 accuracy. Most one-edit pairs differ by a single letter substitution, leaving the other letters in the same positions and therefore much of the rendered image unchanged. DINOv2 and MAE show this pattern particularly clearly: they retrieve one-edit words readily, yet retrieve plural and base partners far less often. Thus, their success on one-edit words can be explained largely by pixel- and patch-level similarity rather than by a general ability to recognize similar spellings. Insertions and deletions provide less visual overlap because they change the word width and, with centered text, shift the remaining letters (\appref{app:withinitem}).

Sensitivity to word form also does not predict semantic retrieval. SigLIP2-B achieves 100\% top-10 accuracy for plural or base partners but only 18\% for capitals, compared with 60\% capital retrieval in DINOv3-H. Conversely, the Qwen3 text encoders achieve 69--74\% synonym accuracy but only 21--26\% one-edit accuracy. Thus, strong retrieval of spelling-related words is neither necessary nor sufficient for strong semantic retrieval. We therefore interpret semantic structure relative to the untrained ViT-B baseline, which captures similarity arising from visual form alone.

%% file: tables/table1_models.tex
\definecolor{tabband}{gray}{0.94}
\begingroup
\def\tablebody{%
\toprule
 & \multicolumn{2}{c}{Neighbor purity} & \multicolumn{4}{c}{Top-10 accuracy} \\
\cmidrule(lr){2-3}\cmidrule(l){4-7}
Model & Nouns & Brands & Capital & Synonym & One-edit word & Plural or base \\
\midrule
\rowcolor{tabband}DINOv3-B & 0.194\,{\scriptsize$\pm$.012} & 0.184\,{\scriptsize$\pm$.016} & 0.000\,{\scriptsize$\pm$.000} & 0.000\,{\scriptsize$\pm$.000} & \underline{0.653}\,{\scriptsize$\pm$.039} & \underline{0.333}\,{\scriptsize$\pm$.054} \\
\rowcolor{tabband}DINOv3-L & \underline{0.223}\,{\scriptsize$\pm$.014} & 0.235\,{\scriptsize$\pm$.023} & \underline{0.117}\,{\scriptsize$\pm$.041} & 0.013\,{\scriptsize$\pm$.009} & \underline{0.820}\,{\scriptsize$\pm$.031} & \underline{0.733}\,{\scriptsize$\pm$.051} \\
\rowcolor{tabband}DINOv3-H & \underline{0.269}\,{\scriptsize$\pm$.017} & \underline{\textbf{0.351}}\,{\scriptsize$\pm$.027} & \underline{\textbf{0.600}}\,{\scriptsize$\pm$.063} & \underline{0.060}\,{\scriptsize$\pm$.019} & \underline{0.607}\,{\scriptsize$\pm$.040} & \underline{0.813}\,{\scriptsize$\pm$.045} \\
\rowcolor{tabband}DINOv3-7B & \underline{0.308}\,{\scriptsize$\pm$.020} & \underline{0.334}\,{\scriptsize$\pm$.031} & \underline{\textbf{0.600}}\,{\scriptsize$\pm$.063} & \underline{0.100}\,{\scriptsize$\pm$.024} & \underline{0.620}\,{\scriptsize$\pm$.040} & \underline{0.853}\,{\scriptsize$\pm$.041} \\
\addlinespace
Web-DINO-1B & 0.187\,{\scriptsize$\pm$.012} & 0.167\,{\scriptsize$\pm$.016} & 0.000\,{\scriptsize$\pm$.000} & 0.007\,{\scriptsize$\pm$.007} & \underline{0.547}\,{\scriptsize$\pm$.041} & \underline{0.293}\,{\scriptsize$\pm$.053} \\
\addlinespace
\rowcolor{tabband}DINOv2-B & 0.179\,{\scriptsize$\pm$.012} & 0.164\,{\scriptsize$\pm$.015} & 0.000\,{\scriptsize$\pm$.000} & 0.000\,{\scriptsize$\pm$.000} & 0.353\,{\scriptsize$\pm$.039} & 0.067\,{\scriptsize$\pm$.029} \\
\rowcolor{tabband}DINOv2-L & 0.174\,{\scriptsize$\pm$.011} & 0.167\,{\scriptsize$\pm$.015} & 0.017\,{\scriptsize$\pm$.017} & 0.007\,{\scriptsize$\pm$.007} & 0.440\,{\scriptsize$\pm$.041} & 0.093\,{\scriptsize$\pm$.034} \\
\rowcolor{tabband}DINOv2-G & 0.181\,{\scriptsize$\pm$.011} & 0.173\,{\scriptsize$\pm$.014} & 0.000\,{\scriptsize$\pm$.000} & 0.013\,{\scriptsize$\pm$.009} & 0.367\,{\scriptsize$\pm$.039} & 0.080\,{\scriptsize$\pm$.031} \\
\addlinespace
MAE-B & 0.182\,{\scriptsize$\pm$.011} & 0.182\,{\scriptsize$\pm$.015} & 0.000\,{\scriptsize$\pm$.000} & 0.000\,{\scriptsize$\pm$.000} & \underline{0.507}\,{\scriptsize$\pm$.041} & 0.040\,{\scriptsize$\pm$.023} \\
MAE-L & 0.186\,{\scriptsize$\pm$.012} & 0.188\,{\scriptsize$\pm$.014} & 0.000\,{\scriptsize$\pm$.000} & 0.000\,{\scriptsize$\pm$.000} & \underline{0.547}\,{\scriptsize$\pm$.041} & 0.080\,{\scriptsize$\pm$.031} \\
MAE-H & 0.180\,{\scriptsize$\pm$.012} & 0.183\,{\scriptsize$\pm$.014} & 0.017\,{\scriptsize$\pm$.017} & 0.007\,{\scriptsize$\pm$.007} & \underline{0.553}\,{\scriptsize$\pm$.041} & 0.093\,{\scriptsize$\pm$.034} \\
\addlinespace
\rowcolor{tabband}CLIP-B & \underline{0.297}\,{\scriptsize$\pm$.019} & \underline{0.343}\,{\scriptsize$\pm$.027} & \underline{0.550}\,{\scriptsize$\pm$.064} & \underline{\textbf{0.200}}\,{\scriptsize$\pm$.033} & \underline{0.853}\,{\scriptsize$\pm$.029} & \underline{0.987}\,{\scriptsize$\pm$.013} \\
\rowcolor{tabband}SigLIP2-B & \underline{\textbf{0.318}}\,{\scriptsize$\pm$.019} & \underline{0.299}\,{\scriptsize$\pm$.025} & \underline{0.183}\,{\scriptsize$\pm$.050} & \underline{0.060}\,{\scriptsize$\pm$.019} & \underline{\textbf{0.947}}\,{\scriptsize$\pm$.018} & \underline{\textbf{1.000}}\,{\scriptsize$\pm$.000} \\
\addlinespace
Untrained ViT-B & 0.183\,{\scriptsize$\pm$.011} & 0.183\,{\scriptsize$\pm$.016} & 0.017\,{\scriptsize$\pm$.017} & 0.007\,{\scriptsize$\pm$.007} & 0.333\,{\scriptsize$\pm$.038} & 0.053\,{\scriptsize$\pm$.026} \\
\midrule
\rowcolor{tabband}Qwen3-0.6B & 0.709\,{\scriptsize$\pm$.022} & 0.632\,{\scriptsize$\pm$.031} & 0.600\,{\scriptsize$\pm$.063} & 0.687\,{\scriptsize$\pm$.038} & 0.213\,{\scriptsize$\pm$.033} & 0.893\,{\scriptsize$\pm$.036} \\
\rowcolor{tabband}Qwen3-4B & 0.711\,{\scriptsize$\pm$.021} & 0.746\,{\scriptsize$\pm$.028} & 0.983\,{\scriptsize$\pm$.017} & 0.733\,{\scriptsize$\pm$.036} & 0.260\,{\scriptsize$\pm$.036} & 0.893\,{\scriptsize$\pm$.036} \\
\rowcolor{tabband}Qwen3-8B & 0.711\,{\scriptsize$\pm$.021} & 0.743\,{\scriptsize$\pm$.027} & 0.967\,{\scriptsize$\pm$.023} & 0.740\,{\scriptsize$\pm$.036} & 0.240\,{\scriptsize$\pm$.035} & 0.867\,{\scriptsize$\pm$.039} \\
\midrule
w2v-BERT~2.0 & 0.201\,{\scriptsize$\pm$.012} & 0.178\,{\scriptsize$\pm$.017} & 0.000\,{\scriptsize$\pm$.000} & 0.013\,{\scriptsize$\pm$.009} & 0.453\,{\scriptsize$\pm$.041} & 0.240\,{\scriptsize$\pm$.049} \\
\midrule
Chance & 0.164\hphantom{\,{\scriptsize$\pm$.000}} & 0.162\hphantom{\,{\scriptsize$\pm$.000}} & 0.006\hphantom{\,{\scriptsize$\pm$.000}} & 0.006\hphantom{\,{\scriptsize$\pm$.000}} & 0.006\hphantom{\,{\scriptsize$\pm$.000}} & 0.006\hphantom{\,{\scriptsize$\pm$.000}} \\
\bottomrule
}%
\setlength{\tabcolsep}{0pt}%
\sbox0{\begin{tabular}{lrrrrrr}\tablebody\end{tabular}}%
\ifdim\wd0<\linewidth%
\setlength{\tabcolsep}{\dimexpr(\linewidth-\wd0)/\numexpr2*7\relax\relax}%
\else\setlength{\tabcolsep}{0pt}\fi%
\begin{tabular}{lrrrrrr}\tablebody\end{tabular}
\endgroup

%% file: sections/5_word_concept_alignment.tex
\begin{table}[!b]
\centering
\footnotesize
\setlength{\defaultaddspace}{2pt}
\renewcommand{\arraystretch}{0.85}
\setlength{\aboverulesep}{0.3ex}
\setlength{\belowrulesep}{0.3ex}
\setlength{\abovecaptionskip}{5pt}
\setlength{\belowcaptionskip}{\abovecaptionskip}
\setlength{\abovecaptionskip}{0pt}
\caption{\textbf{Word directions in DINOv3-H and DINOv3-7B align with the classes they name.} Matched cosine similarity is the mean cosine similarity of the 200 word directions with the visual directions of the classes they name. Word-to-class ranks the 1,000 visual directions for a word direction; class-to-word ranks the 200 word directions for a named class; top-10 accuracy is the share of queries whose target is among the ten nearest, and median rank is the median of the target's rank over the queries, so a median rank of one means the target is nearest for at least half of them.}
\label{tab:concepts}
\input{tables/table2_concepts}
\end{table}

\section{Word-Class Alignment}
\label{sec:alignment}

We test whether word directions $\tau_w$ align with their corresponding visual class directions $\nu_w$ by measuring cosine similarity in the same space. Results are in \tabref{tab:concepts}; additional experiments and ablations are in \appref{app:concept_tables}--\ref{app:composition}.

\paragraph{Named classes.} DINOv3-H and DINOv3-7B show substantial word--class alignment; DINOv3-H is on par with CLIP-B despite no alignment signal during training. Its word direction ranks the named class first among 1,000 for over half of the 200 class names (median rank $=1$). This alignment persists when visual directions are estimated only from images with no detected text ($0.230$ versus $0.234$ with all images; Appendix D.3). SigLIP2-B also aligns, and DINOv3-L weakly. The remaining models show no significant matched alignment despite retrieving one-edit words.

\paragraph{Model size.} DINOv3-H exceeds DINOv3-7B on matched cosine similarity and on retrieval by a factor of 1.7 to 2.0, although DINOv3-7B is no weaker in \tabref{tab:models} and DINOv3-H was distilled from it \citep{simeoni2025dinov3}. We cannot explain this gap. Web-DINO-1B and DINOv2-G, at a billion parameters or more, show no word-class alignments, so size alone does not determine model behavior.

%% file: tables/table2_concepts.tex
\definecolor{tabband}{gray}{0.94}
\begingroup
\def\tablebody{%
\toprule
 & & \multicolumn{2}{c}{Word to class} & \multicolumn{2}{c}{Class to word} \\
\cmidrule(lr){3-4}\cmidrule(l){5-6}
Model & Matched cosine similarity & \shortstack{Top-10\\accuracy} & \shortstack{Median\\rank} & \shortstack{Top-10\\accuracy} & \shortstack{Median\\rank} \\
\midrule
\rowcolor{tabband}DINOv3-B & 0.002\,{\scriptsize$\pm$.003} & 0.010\,{\scriptsize$\pm$.007} & 486 & 0.060\,{\scriptsize$\pm$.017} & 94.5 \\
\rowcolor{tabband}DINOv3-L & \underline{0.025}\,{\scriptsize$\pm$.004} & \underline{0.120}\,{\scriptsize$\pm$.023} & 335 & \underline{0.180}\,{\scriptsize$\pm$.027} & 68 \\
\rowcolor{tabband}DINOv3-H & \underline{0.234}\,{\scriptsize$\pm$.014} & \underline{0.645}\,{\scriptsize$\pm$.034} & \textbf{1} & \underline{0.680}\,{\scriptsize$\pm$.033} & \textbf{1} \\
\rowcolor{tabband}DINOv3-7B & \underline{0.120}\,{\scriptsize$\pm$.012} & \underline{0.325}\,{\scriptsize$\pm$.033} & 154 & \underline{0.405}\,{\scriptsize$\pm$.035} & 32.5 \\
\addlinespace
Web-DINO-1B & 0.001\,{\scriptsize$\pm$.002} & 0.020\,{\scriptsize$\pm$.010} & 477 & 0.045\,{\scriptsize$\pm$.015} & 95 \\
\addlinespace
\rowcolor{tabband}DINOv2-B & -0.002\,{\scriptsize$\pm$.003} & 0.010\,{\scriptsize$\pm$.007} & 527.5 & 0.060\,{\scriptsize$\pm$.017} & 103.5 \\
\rowcolor{tabband}DINOv2-L & -0.001\,{\scriptsize$\pm$.002} & 0.000\,{\scriptsize$\pm$.000} & 485 & 0.045\,{\scriptsize$\pm$.015} & 99.5 \\
\rowcolor{tabband}DINOv2-G & 0.001\,{\scriptsize$\pm$.002} & 0.005\,{\scriptsize$\pm$.005} & 463 & 0.045\,{\scriptsize$\pm$.015} & 94 \\
\addlinespace
MAE-B & -0.011\,{\scriptsize$\pm$.006} & 0.005\,{\scriptsize$\pm$.005} & 559 & 0.020\,{\scriptsize$\pm$.010} & 113 \\
MAE-L & 0.003\,{\scriptsize$\pm$.004} & 0.015\,{\scriptsize$\pm$.009} & 491 & 0.045\,{\scriptsize$\pm$.015} & 98.5 \\
MAE-H & 0.002\,{\scriptsize$\pm$.004} & 0.005\,{\scriptsize$\pm$.005} & 461.5 & 0.050\,{\scriptsize$\pm$.015} & 96.5 \\
\addlinespace
\rowcolor{tabband}CLIP-B & \underline{\textbf{0.237}}\,{\scriptsize$\pm$.010} & \underline{\textbf{0.855}}\,{\scriptsize$\pm$.025} & \textbf{1} & \underline{\textbf{0.845}}\,{\scriptsize$\pm$.026} & \textbf{1} \\
\rowcolor{tabband}SigLIP2-B & \underline{0.184}\,{\scriptsize$\pm$.009} & \underline{0.630}\,{\scriptsize$\pm$.034} & 2 & \underline{0.735}\,{\scriptsize$\pm$.031} & 2 \\
\addlinespace
Untrained ViT-B & 0.005\,{\scriptsize$\pm$.008} & 0.005\,{\scriptsize$\pm$.005} & 511.5 & 0.045\,{\scriptsize$\pm$.015} & 99 \\
\midrule
Chance & 0.000\hphantom{\,{\scriptsize$\pm$.000}} & 0.010\hphantom{\,{\scriptsize$\pm$.000}} &  & 0.050\hphantom{\,{\scriptsize$\pm$.000}} &  \\
\bottomrule
}%
\setlength{\tabcolsep}{0pt}%
\sbox0{\begin{tabular}{lrrrrr}\tablebody\end{tabular}}%
\ifdim\wd0<\linewidth%
\setlength{\tabcolsep}{\dimexpr(\linewidth-\wd0)/\numexpr2*6\relax\relax}%
\else\setlength{\tabcolsep}{0pt}\fi%
\begin{tabular}{lrrrrr}\tablebody\end{tabular}
\endgroup

%% file: sections/6_finetuning_and_functional_use.tex
\section{Alignment and Finetuning}
\label{sec:encoders}

\begin{figure}[t]
\centering
\includegraphics[width=\textwidth]{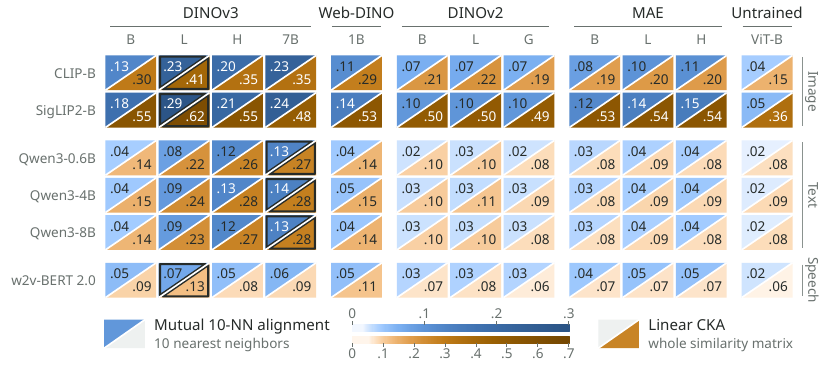}
\caption{\textbf{DINOv3-H and DINOv3-7B are most aligned with text encoders.}
Each cell compares a column model and row encoder on the lexical vocabulary; the image rows, CLIP-B and SigLIP2-B, receive rendered word images. The upper triangle shows mutual 10-NN alignment (shared nearest neighbors; chance 0.006), and the lower triangle shows linear CKA between centered similarity matrices (chance 0.005--0.052). Each measure uses a logarithmic scale, outlines mark row maxima, and every cell exceeds a null that relabels words of equal length (\appref{app:statistics}).}
\label{fig:language_alignment}
\end{figure}

We now compare each model with text encoders, given the 1,561 words, names, and pseudowords as text, and with a speech encoder, given them as audio (Section~\ref{sec:cross_encoder}; results in~\figref{fig:language_alignment}).

\paragraph{Text encoders.} DINOv3-H and DINOv3-7B share 12--14\% of their ten nearest neighbors with the text encoders, the highest agreement among the self-supervised vision models. Linear CKA shows the same pattern. Although text encoders never see the rendered words, their representations retain spelling information: Qwen3 encoders retrieve 21--26\% of one-edit partners (\tabref{tab:models}). Vision models without detectable semantic structure share at most 5\% of their neighbors with text encoders, suggesting that word form contributes to agreement but does not fully explain it.

\paragraph{Shared word form.} CLIP-B, SigLIP2-B, and the speech encoder order the models differently. Every vision-only model agrees more with CLIP-B and SigLIP2-B than with any text encoder, and MAE-H reaches a linear CKA of 0.54 with SigLIP2-B, above DINOv3-7B's 0.48, although MAE shows no semantic structure. Two image encoders see the same rendered letters, and a word's sound follows its spelling, so we read this agreement as largely reflecting word form. DINOv3-L, the self-supervised model that finds the most one-edit words, has the largest values against all three, yet shares fewer nearest neighbors with every text encoder than DINOv3-H or DINOv3-7B.

\paragraph{Finetuning.} Contrastive finetuning on images with no detected text improves synonym retrieval and noun grouping in DINOv3-H, while the other nine finetuned models change minimally (\appref{app:finetune_recipe}). Using four rendered images per word, synonym top-10 accuracy rises from about 2\% to 11\% with text supervision and 8\% with speech supervision, averaged over five typography draws. After text finetuning, both synonym retrieval and noun purity exceed the pretrained model's scores using all 96 images. Thus, caption supervision makes it easier for DINOv3-H to recover its lexical structure from fewer rendered examples, without yielding comparable gains in the other models.

%% file: sections/7_discussion.tex
\section{Discussion}
\paragraph{The scope and possible sources of semantic knowledge.}
DINOv3-H and DINOv3-7B exhibit lexical semantic structure connecting written words to one another and to depicted object categories.
Not all relations are equally strong: Category membership and country-capital associations are more clearly expressed than synonymy.
Similarity and association are distinct \citep{hill2015simlex}, so factual relationships do not imply equally strong synonym knowledge. These findings do not establish taxonomic hierarchies or sentence-level composition. Nor do consistent offsets improve analogy retrieval over the country-only baseline, reinforcing the distinction between geometric regularity and analogical reasoning \citep{linzen2016issues}.

The source of this structure remains unresolved. Words can acquire associations from surrounding visual content or through text embedded in frames; the latter could provide linguistic co-occurrence signals even for semantic relations not directly instantiated in the depicted scene. Embedded text influences CLIP training \citep{lin2023parrot}; it is also a potential source of semantic information in DINOv3's images, which are drawn mainly from public Instagram posts \citep{simeoni2025dinov3}. 
Image-only training therefore need not mean learning exclusively from depicted, concrete objects; vision models' knowledge of abstract semantics remains an open question. 

\paragraph{What finetuning changes.} Finetuning improves synonym retrieval and noun purity in DINOv3-H, with little change in the other finetuned models. After text finetuning, just four rendered images per word yield higher scores on both measures than all 96 images before finetuning. This is consistent with a partial functional-dormancy interpretation: caption supervision makes the existing lexical structure easier to recover. However, these results do not distinguish eliciting existing relations from learning new ones. The interpretation connects to work on eliciting latent capabilities \citep{NEURIPS2025_dd16e008}, although the evidence here concerns representations rather than downstream behavior. Claims of representational convergence also remain limited by the cross-encoder null, which controls for word length but not all shared properties of word form.

\paragraph{Implications for downstream systems and safety.}
Our results demonstrate that the absence of language supervision does not guarantee that written text is inert to systems built on a vision encoder. Image-only pretraining can produce lexical and factual associations, and later finetuning toward language can strengthen some of them. Harmful requests rendered in images can bypass safeguards in vision-language systems \citep{gong2025figstep}. If scaling broadens the semantic knowledge of image-only pretrained encoders, evaluations should examine both emerging capabilities and their downstream use \citep{shevlane2023model}. Our experiments establish neither instruction following nor harmful behavior, but motivate testing whether rendered instructions influence downstream systems.

%% file: sections/8_conclusion.tex
\section{Conclusion}
We show that vision transformers pretrained without explicit language supervision can encode
semantic relations among written words. In DINOv3-H and DINOv3-7B, word directions capture
category membership and country-capital associations under spelling controls, with weaker evidence
for synonymy. These directions also align with representations of the depicted object categories
they name and share neighborhood structure with text encoders. Contrastive finetuning on images
screened for text raises synonym retrieval and the grouping of nouns by category in DINOv3-H,
lowers its capital retrieval, and changes the other finetuned models little. These findings establish
word-level semantic structure, rather than sentence understanding or instruction following. Image-
only pretraining can therefore teach vision models relationships among written words and their visual
referents, extending what they learn beyond the appearance of text.

%% file: appendices/A_data_and_rendering.tex
\section{Data and Rendering}
\label{app:data}

\paragraph{Word selection.} We sampled the vocabulary mainly from WordNet and chose round list sizes. Eight lists record the words, their roles within an item, and their category and spelling-cluster labels (\tabref{tab:wordlists}). We checked for repeated words, rendering fit, morphological relations, letter overlap, and equal category counts within spelling clusters.

The construction records also describe language-model proposals for synonym pairs and antonym and negation triples, followed by screening. One-edit words were found by searching possible letter edits. Noun candidates combined a WordNet pool with language-model proposals. Separate procedures formed plural pairs, filtered a country-capital list, and generated pseudowords.

\paragraph{Word frequency.} We estimated English word frequencies using \texttt{wordfreq}, aiming for at least one occurrence per million words. The estimates do not distinguish different meanings of a word. The final lists include 15 capital names, 152 nouns, and 121 brand names below this target. Within three-word items, the median ratio of highest to lowest frequency is 6.31 in the synonym and one-edit list, 8.93 in the synonym and antonym list, and 21.38 in the negation list.

\paragraph{Spelling within items.} All 150 one-edit words have Levenshtein edit distance one from their anchor: 135 substitutions, seven insertions, and eight deletions. Synonyms in this list are on average 1.23 letters longer than the anchor. A frequency above one per million does not by itself establish that a one-edit word is an ordinary dictionary word; the list includes \emph{ain}, \emph{lan}, and \emph{icc}.

\paragraph{Letter overlap.} We measure overlap as the number of distinct letters shared by two words divided by the number of distinct letters in either word. Repeated letters are counted once. For each meaning relation, we compare the average overlap with that of random word pairs matched in length. The largest excess is 0.0125. This is an average for a relation, so individual pairs may share more letters. We also use WordNet relations and spelling rules to check for inflectional and derivational relatives. All meaning pairs pass these checks.

\begin{tablock}
\centering
\footnotesize
\setlength{\belowcaptionskip}{\abovecaptionskip}
\setlength{\abovecaptionskip}{0pt}
\captionof{table}{The eight vocabulary lists. Items are triples in the first three lists, pairs in the next two, and individual words or pseudowords in the remaining lists. Each triple defines two pairs sharing an anchor. The Words column includes names and pseudowords. Each appears in one list only. The Pairs column counts pairs before the query direction is reversed.}
\label{tab:wordlists}
\begin{tabular}{@{}p{0.24\linewidth}rrrp{0.35\linewidth}@{}}
\toprule
List & Items & Words & Pairs & Example and purpose \\
\midrule
Synonym and one-edit comparison & 150 & 450 & 300 & \emph{torch--flashlight}, \emph{torch--touch}: meaning and spelling \\
Synonym and antonym & 50 & 150 & 100 & \emph{brave--bold}, \emph{brave--cowardly}: synonymy and opposition \\
Synonym and apparent negation/base & 35 & 105 & 70 & \emph{disappear--vanish}, \emph{disappear--appear}: meaning and similar form \\
Singular and plural & 40 & 80 & 40 & \emph{road--roads}: form with related meaning \\
Country and capital & 60 & 120 & 60 & \emph{germany--berlin}: factual association \\
Common nouns & 384 & 384 & --- & Category membership \\
Brands & 192 & 192 & --- & Industry membership \\
Pseudowords & 80 & 80 & --- & Retrieval candidates and centering \\
\midrule
Total & --- & 1,561 & 570 & \\
\bottomrule
\end{tabular}
\end{tablock}

\paragraph{Pairs.} Each three-word item defines two pairs with the same anchor. For example, \emph{torch}, \emph{flashlight}, and \emph{touch} define the pairs \emph{torch--flashlight} and \emph{torch--touch}. When \emph{flashlight} is the retrieval target, \emph{touch} remains among the candidates. The five relation lists contain 570 pairs before the query direction is reversed. The directions used in each analysis are specified in \appref{app:measures}. Synonym pairs occur in all three lists of three-word items.

The relation lists contain 905 distinct words and names. No word, name, or pseudoword appears twice across the eight lists. Names are lowercased for rendering. Forty-three of the separate 200 class names also occur in the first vocabulary. Their rendered images are identical, but their word directions differ because they are centered over different vocabularies.

\paragraph{Exceptions to the relation labels.} Some words in the negation list only resemble negations or share a stem with the recorded base. Examples include \emph{injury--jury}, \emph{invent--vent}, \emph{distract--tract}, and \emph{impending--pending}. The pair \emph{harmless--harmful} changes a suffix, and \emph{invaluable} does not simply negate \emph{valuable}. The full list, therefore, compares synonyms with words of similar spelling but is not a uniform test of negation. The plural list also contains exceptions. The pair \emph{ala--alas} is not a standard singular--plural pair, and some plurals have distinct lexicalized senses.

\paragraph{Categories and spelling clusters.} The noun categories are animal, device, food, person, plant, and structure, with 64 words each. The brand industries are beverage, carmaker, electronics, internet and software, retail chain, and video game, each with 32 names. Each list has four spelling clusters. Every noun category has 16 words in each cluster, and every brand industry has eight. Category and spelling-cluster labels therefore have Cram\'er's $V=0$. This balance does not remove associations between category and individual spelling features, such as suffixes or word length.

The recorded noun-clustering procedure represents each word by TF--IDF-weighted character sequences of two to four letters, including word boundaries. It reduces the vectors to 50 dimensions using truncated singular value decomposition, normalizes them, and fits four k-means centers. Words are then assigned to the centers, with an equal number from each semantic category in each cluster. A word's assignment therefore depends on its spelling, the other candidate words, and the required category counts.

\paragraph{Pseudowords.} We generated 80 four- to eight-letter pseudowords from consonant onsets, vowel sequences, and endings, filtering out \texttt{wordfreq} and WordNet entries, given names, and close spelling matches to specified words or accepted pseudowords. They serve as additional retrieval candidates, contribute to vocabulary centering, and are included in cross-encoder comparisons, but are neither queried nor scored as correct partners in word-pair tests. They are excluded from noun and brand candidate sets. Scrambled images, not pseudowords, provide the matched visual controls.

\paragraph{Background images.} We use 96 ImageNet-1k validation photographs from 96 different classes. We screen candidates for image variation and central clutter, and rank them using CLIP-B prompts that describe image quality. We screen for text using CRAFT \citep{baek2019craft}, then manually inspect every selected background to confirm that it contains no visible text.

\paragraph{Typography.} Every word, name, and pseudoword fits inside the central 224-pixel crop-safe window of a 256-pixel image at every font size. The widest measures approximately 187 pixels. We pair each of the six fonts with each of the four vertical position bands, yielding 24 settings. Each font size and color occurs equally often, and half the settings use a backing panel. Each setting is used on four distinct backgrounds. We render all words, names, and pseudowords on these same 96 backgrounds with the assigned typography. Each background has one setting, rather than all 24. The text center is fixed on each background, while the text width and panel width vary with the word.

The fonts are DejaVu Sans, PT Sans, DejaVu Serif, Nimbus Roman, DejaVu Sans Mono, and URW Gothic. Small, medium, and large text use font sizes of 13, 20, and 28 pixels. The colors are near-black, near-white, red, blue, green, and purple. Text is centered horizontally. Its vertical center lies within one of four equal bands of the crop-safe window. The backing is bare or a translucent panel, dark behind near-white text and white behind the other colors. Five of the ten pairs of typographic factors are exactly balanced. The strongest association among the others is between font and color, with Cramér's V = 0.32. Rendering the 1,481 words and names and 80 pseudowords on 96 backgrounds, each with a word image and control image, plus the 96 bare backgrounds, gives 299,808 images.

\paragraph{Control image.} We replace the letters' opacity mask inside the text bounding box with a phase-scrambled mask. We keep the constant Fourier term, draw the other phases uniformly from $[-\pi,\pi]$, take the real part of the inverse transform, rescale to the original mask's mean and standard deviation, and clip to $[0,1]$. Word and control images share the background, bounding box, position, color, and panel. Taking the real part and clipping can change the Fourier amplitudes, opacity sum, and variance. The subtraction therefore compares intact and scrambled lettering while allowing some differences in visual form to remain. The phases are fixed for each word or pseudoword on each background.

\paragraph{Centering.} Before centering, the squared norm of the vocabulary mean of the uncentered difference vectors accounts for a median of 88\% of their mean squared norm across the fourteen models, ranging from 27\% in CLIP-B to 93\% in DINOv2-G. We subtract this mean from $\hat{\tau}_w$ to obtain the word direction $\tau_w$ in \eqref{eq:tau}. A word's direction therefore depends on the rest of the vocabulary. For lexical and cross-encoder comparisons, the mean includes 1,481 words and names and 80 pseudowords. For word--class comparisons, it includes only the 200 class names.

\begin{figure}[!htb]
\centering
\includegraphics[width=\textwidth]{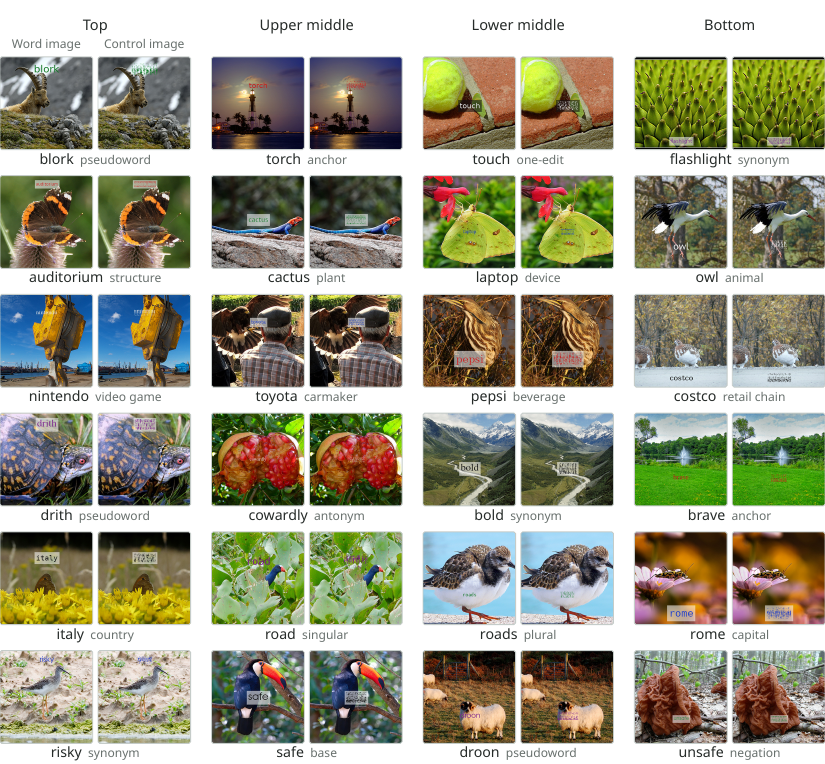}
\caption{\textbf{Word images and control images.} Each row shows one font and each column one vertical position band, covering the 24 typography settings. Each cell shows a word image beside its control image. The control uses the same background, typography, and text color, with the letters' opacity mask replaced by a phase-scrambled copy. The words and pseudowords are examples from the eight lists in \tabref{tab:wordlists}, with their roles indicated.}
\label{fig:rendering}
\end{figure}

%% file: appendices/B_measures.tex
\section{Measures and Statistical Tests}
\label{app:statistics}
\label{app:alignment_tests}

We define the measures used to assess lexical structure and alignment, then describe the statistical tests, uncertainty estimates, and comparisons with the untrained baseline.

\subsection{Measures}
\label{app:measures}

\paragraph{Neighbors and retrieval.} We rank candidates by cosine similarity, excluding the query itself. For neighbor purity, we find each noun's ten nearest neighbors among the other nouns and each brand's among the other brands. Purity is the fraction that share the query's category or industry, averaged across queries. Partner retrieval searches the full vocabulary for a word's designated partner. We report top-10 accuracy, the fraction of queries whose partner ranks among the ten nearest candidates, and median rank across queries, resolving ties in the target's favor. The main lexical results use the first word of each pair as the query. We report both directions separately in \appref{app:lexical_retrieval} and pool them for the finetuning analysis (\secref{sec:encoders}).

\paragraph{Word and visual directions.} Three measures compare the word directions of the 200 class names with the visual directions. Word-to-class retrieval queries with the word direction of each class name and ranks the 1,000 visual directions, so top-10 chance is 0.010. Class-to-word retrieval runs the other way, from the visual direction of each of the 200 named classes to the 200 word directions, so top-10 chance is 0.050. Matched cosine similarity ranks no candidates and instead averages, over the 200 class names, the cosine similarity between a name's word direction and the visual direction of the class it names, both taken from the same model and layer. Its chance is zero, and its null reassigns classes to words (\appref{app:tests}).

\paragraph{Alignment across encoders.} Mutual 10-NN alignment and linear CKA compare two models or encoders through the cosine similarities each gives the same 1,561 words, names, and pseudowords, since their vectors share no basis and often differ in width. Every model and encoder $E$ gives a matrix $S_E$ of cosine similarities between its centered vectors, and $N^E_{10}(w)$ is the set of the ten other words nearest to $w$ under $S_E$. Mutual 10-NN alignment \citep{huh2024platonic} between two encoders $A$ and $B$ is
\begin{equation}
\label{eq:mutual_knn}
\frac{1}{n}\sum_{w} \frac{\bigl|N^A_{10}(w) \cap N^B_{10}(w)\bigr|}{10},
\qquad n = 1{,}561,
\end{equation}
where the sum over $w$ covers the full vocabulary. The score is symmetric in $A$ and $B$, with an expected value of 0.006 when the two neighborhoods are unrelated. Linear CKA \citep{kornblith2019cka} centers each matrix over the vocabulary, $K_E = H S_E H$ with $H = I - \tfrac{1}{n}\mathbf{1}\mathbf{1}^{\top}$. Here, $H$ is the centering matrix, $I$ is the $n \times n$ identity matrix, and $\mathbf{1}$ is a column vector of $n$ ones. The score is
\begin{equation}
\label{eq:cka}
\frac{\langle K_A, K_B\rangle_F}{\|K_A\|_F\,\|K_B\|_F},
\end{equation}
Here, $\langle \cdot, \cdot \rangle_F$ is the Frobenius inner product, the sum of products of corresponding matrix entries. The Frobenius norm $\|\cdot\|_F$ is the square root of the sum of squared entries. Linear CKA is one when $K_A$ and $K_B$ are proportional. As in \citet{kornblith2019cka}, we keep the diagonal of each matrix, each word's similarity with itself. When one encoder's words are relabeled uniformly at random, linear CKA has expectation $\operatorname{tr} K_A \, \operatorname{tr} K_B / \bigl((n-1)\,\|K_A\|_F\,\|K_B\|_F\bigr)$, because the rows and columns of a centered matrix sum to zero. The trace $\operatorname{tr}$ is the sum of the diagonal entries. That chance depends on the two matrices and is 0.005 to 0.052 across the cells of \figref{fig:language_alignment}. The two measures weigh different things: mutual 10-NN alignment depends only on each word's ten nearest neighbors, and linear CKA on all the similarities, weighted toward the principal components of largest variance.

\subsection{Statistical tests}
\label{app:tests}

\paragraph{Permutation tests and corrections.} Our permutation tests keep the vectors fixed and shuffle the labels or assignments relevant to each measure. For matched cosine similarity, we randomly reassign classes to words 2,000 times. For category structure, we shuffle category labels among words of similar length (\appref{app:category}). With $R$ random permutations, the one-sided $p$-value is $(r+1)/(R+1)$, where $r$ counts scores at least as large as the observed score. The two-clue analysis instead evaluates every combination of sign flips across the twelve clue grids, using a two-sided test for the residual and one-sided tests for the other outcomes (\appref{app:composition}). Tables report uncorrected $p$-values. Where indicated, we apply Holm correction separately across all fourteen models in the within-item comparisons and across the thirteen pretrained models in the two-clue analysis.

\paragraph{Null for alignment across encoders.} To control for word length, we shuffle word labels among words of equal length in one encoder while keeping the other encoder's similarity matrix fixed. We recompute both alignment measures for 1,000 permutations, using the same permutations for every comparison in \figref{fig:language_alignment}. For every comparison, including those involving the untrained ViT-B, the observed score exceeds all corresponding permutation scores. All comparisons remain significant after Benjamini--Hochberg correction within each measure.

\paragraph{Standard errors and intervals.} Standard errors reflect variation across words, classes, pairs, or items, with images, typography, and the held-out split fixed. We use binomial standard errors for top-10 and nearest-centroid accuracy. For item-level similarities, we divide the standard deviation across items by the square root of the item count. For neighbor purity and matched cosine similarity, we recompute scores on 1,000 random subsets of 80\% of the words, sampled without replacement. For purity, neighbors are drawn only from the selected subset. We multiply the standard deviation across subsets by $\sqrt{m(n-1)/(n(n-m))}$, where $n$ is the full sample size and $m=\lfloor 0.8n\rfloor$, to estimate the full-sample standard error. For the two-clue analysis, we report 95\% percentile bootstrap intervals from 20,000 resamples of the twelve clue grids.

\paragraph{Comparisons with the untrained ViT-B.} We underline values that exceed the untrained ViT-B by more than twice the estimated standard error of the difference. We estimate this error as $\sqrt{s^2+s_0^2}$, where $s$ and $s_0$ are the standard errors for the trained and untrained models. The rule does not account for pairing across models or adjust for multiple comparisons. Values without standard errors are not underlined.

%% file: appendices/C_lexical_space_analyses.tex
\section{Lexical Structure}
\label{app:lexical}
\suppressfloats[t]

We examine how query direction and spelling affect word-pair retrieval, test category grouping on held-out words, and trace lexical structure across model depth.

\subsection{Pairs}
\label{app:lexical_retrieval}

\paragraph{Both query directions.} \tabref{tab:lexical_retrieval} reports partner retrieval with either word in a pair used as the query. Reversing the query direction lowers country-capital retrieval in DINOv3-H and DINOv3-7B but raises it in CLIP-B. Synonym retrieval improves in both large DINOv3 models when the partner is used as the query. The main results in \tabref{tab:models} use forward queries only and combine the plural and base pairs in one column.

\begin{table}[p]
\centering
\footnotesize
\setlength{\tabcolsep}{3pt}
\setlength{\belowcaptionskip}{\abovecaptionskip}
\setlength{\abovecaptionskip}{0pt}
\caption{Partner retrieval at the final layer. Top-10 accuracy is the fraction of queries whose partner is among the ten nearest candidates in the full vocabulary, excluding the query. Queries are (a) the first word of each pair (anchor, negation, singular noun, or country) and (b) its partner. Small numbers are standard errors; underlining marks values that exceed the untrained ViT-B by more than twice the standard error of the difference. Bold marks each column's maximum within a panel. Text and speech encoders are excluded from highlighting.}
\label{tab:lexical_retrieval}
\input{tables/tableB5_lexical_retrieval}
\end{table}

\paragraph{Projection in \panelref{fig:nearest}{b}.}\phantomsection\label{app:projection} The panel shows a two-dimensional projection of six country-capital pairs in DINOv3-H, selected after inspecting the full set for familiar names and a legible layout. We normalize the word directions to unit length and center the displayed endpoints. The horizontal axis follows the mean country-to-capital offset of the undisplayed pairs. The vertical axis is the leading principal component of the displayed pairs' midpoints after removing the horizontal component. For display, we stretch the horizontal coordinates by a factor of 1.6 relative to the vertical coordinates.

\paragraph{Comparisons within an item.}\phantomsection\label{app:withinitem} \tabref{tab:items} compares each word's similarity to a synonym and a competing word, keeping the query fixed. The competitor is a one-edit word, an antonym, or the base form of a negation. Two-sided exact sign tests assess whether the synonym is nearer in more or fewer than half the items. All vision models favor one-edit words over synonyms, although synonym similarity is higher in DINOv3-H, DINOv3-7B, CLIP-B, and SigLIP2-B than in the other vision models. No vision model favors synonyms over antonyms after Holm correction. Negated words are usually closer to their base forms than to their synonyms in DINOv3, CLIP-B, SigLIP2-B, and Web-DINO-1B; DINOv2, MAE, and the untrained ViT-B show no clear preference.

DINOv2 and MAE retrieve one-edit words more readily than plurals or base forms (\tabref{tab:lexical_retrieval}). The way we center the words matters for this comparison. Most one-edit words differ by a single letter without changing word length, so much of the text remains in the same position. Plurals and base forms add or remove letters, changing the word's width. Centering then shifts the whole word, including the letters shared by both forms, and reduces their visual overlap. Retrieving one-edit words therefore does not, on its own, demonstrate recognition of similar spelling beyond this overlap.

\begin{tablock}
\centering
\footnotesize
\setlength{\belowcaptionskip}{\abovecaptionskip}
\setlength{\abovecaptionskip}{0pt}
\captionof{table}{Word-pair comparisons at the final layer. Counts give the number of items in which the synonym is closer to the query than the named competitor. Cosine similarities to the anchor are expressed in standard deviations above the vocabulary-wide pairwise mean, with standard errors over items. Top-10 hits count successful synonym retrieval in either direction among 75 concrete and 75 abstract anchors. Stars mark uncorrected $p<0.05$ favoring synonyms (two-sided sign test) or concrete anchors (one-sided Fisher exact test). Underlining marks similarities that exceed the untrained ViT-B by more than twice the standard error of the difference. Bold marks the largest synonym similarity and the largest hit count across both anchor groups. Text and speech encoders are excluded from tests and highlighting.}
\label{tab:items}
\input{tables/tableB3_items}
\end{tablock}

\subsection{Grouping by category}
\label{app:category}

\paragraph{Held-out words.} \tabref{tab:category} reports neighbor purity and nearest-centroid accuracy for nouns and brands. The latter tests whether category grouping extends to held-out words. We split the words evenly within each combination of category and spelling cluster, then average the unit-length word directions in one half to form the centroids. Each word in the other half is assigned to the nearest centroid by cosine similarity. DINOv3-H, DINOv3-7B, CLIP-B, and SigLIP2-B have the highest accuracy and neighbor purity among the vision models for both nouns and brands, followed by DINOv3-L.

\begin{tablock}
\centering
\footnotesize
\setlength{\belowcaptionskip}{\abovecaptionskip}
\setlength{\abovecaptionskip}{0pt}
\captionof{table}{Category grouping at the final layer. Neighbor purity is the fraction of a word's ten nearest neighbors in the noun or brand list that share its category or industry. Nearest-centroid accuracy measures category assignment on held-out words. Permutation tests control for word length (\appref{app:category}); small numbers are standard errors. Underlining marks values that exceed the untrained ViT-B by more than twice the standard error of the difference. Bold marks column maxima. Text and speech encoders are excluded from tests and highlighting.}
\label{tab:category}
\input{tables/tableB4_category}
\end{tablock}

\paragraph{Word form.} The permutation tests in \tabref{tab:category} assess category grouping while accounting for differences in word length across categories. We shuffle category labels within word-length terciles, also preserving spelling clusters for nearest-centroid accuracy. We use 2,000 permutations for each test. The observed scores for DINOv3-H, DINOv3-7B, CLIP-B, and SigLIP2-B exceed every permutation score for both measures on nouns and brands. The untrained ViT-B also shows significant neighbor purity, so we assess the trained models relative to this baseline (\secref{sec:lexical}). Restricting noun neighbors to the query's spelling cluster likewise leaves DINOv3-H and DINOv3-7B above the untrained baseline.

\subsection{Depth}
\label{app:lexical_depth}

\tabref{tab:lexical_depth} reports noun neighbor purity and country-to-capital retrieval at successive depths in each model. Noun purity increases from the middle layers onward in DINOv3-H, DINOv3-7B, CLIP-B, and SigLIP2-B. DINOv3-L separates from the untrained baseline mainly at the final layer, while the other vision models remain close to it throughout. Country-capital retrieval improves later: DINOv3-H and DINOv3-7B rise sharply at the last two sampled depths, while CLIP-B improves at the final layer. These models also begin to rank capitals first at those depths.

\begin{table}[p]
\centering
\footnotesize
\setlength{\belowcaptionskip}{\abovecaptionskip}
\setlength{\abovecaptionskip}{0pt}
\caption{Noun neighbor purity (a) and country-to-capital top-10 accuracy (b) across model depth. Relative depth is the layer index divided by the number of transformer layers; column labels give the mean across models at each sampled depth. The input embedding layer is excluded. Small numbers are standard errors. Underlining marks values that exceed the untrained ViT-B at the corresponding depth by more than twice the standard error of the difference. Bold marks each model's maximum across depths when it exceeds chance.}
\label{tab:lexical_depth}
\input{tables/tableB7_lexical_depth}
\end{table}

%% file: tables/tableB5_lexical_retrieval.tex
\definecolor{tabband}{gray}{0.94}
\begingroup
\def\tablebody{%
\toprule
 & \multicolumn{5}{c}{Meaning} & \multicolumn{3}{c}{Spelling} \\
\cmidrule(lr){2-6}\cmidrule(l){7-9}
Model & Capital & Synonym & \shortstack{Synonym\\(antonym\\list)} & Antonym & \shortstack{Synonym\\(negation\\list)} & \shortstack{One-edit\\word} & Plural & Base \\
\midrule
\multicolumn{9}{l}{\textbf{(a) Forward}} \\
\addlinespace[2pt]
\rowcolor{tabband}DINOv3-B & 0.000\,{\tiny$\pm$.000} & 0.000\,{\tiny$\pm$.000} & 0.000\,{\tiny$\pm$.000} & 0.000\,{\tiny$\pm$.000} & 0.000\,{\tiny$\pm$.000} & \underline{0.653}\,{\tiny$\pm$.039} & \underline{0.450}\,{\tiny$\pm$.079} & 0.200\,{\tiny$\pm$.068} \\
\rowcolor{tabband}DINOv3-L & \underline{0.117}\,{\tiny$\pm$.041} & 0.013\,{\tiny$\pm$.009} & 0.000\,{\tiny$\pm$.000} & 0.020\,{\tiny$\pm$.020} & 0.029\,{\tiny$\pm$.028} & \underline{0.820}\,{\tiny$\pm$.031} & \underline{0.950}\,{\tiny$\pm$.034} & \underline{0.486}\,{\tiny$\pm$.084} \\
\rowcolor{tabband}DINOv3-H & \underline{\textbf{0.600}}\,{\tiny$\pm$.063} & \underline{0.060}\,{\tiny$\pm$.019} & 0.020\,{\tiny$\pm$.020} & 0.000\,{\tiny$\pm$.000} & 0.086\,{\tiny$\pm$.047} & \underline{0.607}\,{\tiny$\pm$.040} & \underline{\textbf{1.000}}\,{\tiny$\pm$.000} & \underline{0.600}\,{\tiny$\pm$.083} \\
\rowcolor{tabband}DINOv3-7B & \underline{\textbf{0.600}}\,{\tiny$\pm$.063} & \underline{0.100}\,{\tiny$\pm$.024} & 0.020\,{\tiny$\pm$.020} & 0.060\,{\tiny$\pm$.034} & 0.086\,{\tiny$\pm$.047} & \underline{0.620}\,{\tiny$\pm$.040} & \underline{\textbf{1.000}}\,{\tiny$\pm$.000} & \underline{0.686}\,{\tiny$\pm$.078} \\
\addlinespace
Web-DINO-1B & 0.000\,{\tiny$\pm$.000} & 0.007\,{\tiny$\pm$.007} & 0.020\,{\tiny$\pm$.020} & 0.000\,{\tiny$\pm$.000} & 0.000\,{\tiny$\pm$.000} & \underline{0.547}\,{\tiny$\pm$.041} & \underline{0.475}\,{\tiny$\pm$.079} & 0.086\,{\tiny$\pm$.047} \\
\addlinespace
\rowcolor{tabband}DINOv2-B & 0.000\,{\tiny$\pm$.000} & 0.000\,{\tiny$\pm$.000} & 0.000\,{\tiny$\pm$.000} & 0.000\,{\tiny$\pm$.000} & 0.029\,{\tiny$\pm$.028} & 0.353\,{\tiny$\pm$.039} & 0.100\,{\tiny$\pm$.047} & 0.029\,{\tiny$\pm$.028} \\
\rowcolor{tabband}DINOv2-L & 0.017\,{\tiny$\pm$.017} & 0.007\,{\tiny$\pm$.007} & 0.020\,{\tiny$\pm$.020} & 0.000\,{\tiny$\pm$.000} & 0.000\,{\tiny$\pm$.000} & 0.440\,{\tiny$\pm$.041} & 0.125\,{\tiny$\pm$.052} & 0.057\,{\tiny$\pm$.039} \\
\rowcolor{tabband}DINOv2-G & 0.000\,{\tiny$\pm$.000} & 0.013\,{\tiny$\pm$.009} & 0.000\,{\tiny$\pm$.000} & 0.000\,{\tiny$\pm$.000} & 0.000\,{\tiny$\pm$.000} & 0.367\,{\tiny$\pm$.039} & 0.125\,{\tiny$\pm$.052} & 0.029\,{\tiny$\pm$.028} \\
\addlinespace
MAE-B & 0.000\,{\tiny$\pm$.000} & 0.000\,{\tiny$\pm$.000} & 0.020\,{\tiny$\pm$.020} & 0.020\,{\tiny$\pm$.020} & 0.000\,{\tiny$\pm$.000} & \underline{0.507}\,{\tiny$\pm$.041} & 0.050\,{\tiny$\pm$.034} & 0.029\,{\tiny$\pm$.028} \\
MAE-L & 0.000\,{\tiny$\pm$.000} & 0.000\,{\tiny$\pm$.000} & 0.000\,{\tiny$\pm$.000} & 0.000\,{\tiny$\pm$.000} & 0.029\,{\tiny$\pm$.028} & \underline{0.547}\,{\tiny$\pm$.041} & 0.125\,{\tiny$\pm$.052} & 0.029\,{\tiny$\pm$.028} \\
MAE-H & 0.017\,{\tiny$\pm$.017} & 0.007\,{\tiny$\pm$.007} & 0.000\,{\tiny$\pm$.000} & 0.000\,{\tiny$\pm$.000} & 0.029\,{\tiny$\pm$.028} & \underline{0.553}\,{\tiny$\pm$.041} & 0.125\,{\tiny$\pm$.052} & 0.057\,{\tiny$\pm$.039} \\
\addlinespace
\rowcolor{tabband}CLIP-B & \underline{0.550}\,{\tiny$\pm$.064} & \underline{\textbf{0.200}}\,{\tiny$\pm$.033} & \underline{\textbf{0.200}}\,{\tiny$\pm$.057} & \textbf{0.120}\,{\tiny$\pm$.046} & \underline{\textbf{0.143}}\,{\tiny$\pm$.059} & \underline{0.853}\,{\tiny$\pm$.029} & \underline{\textbf{1.000}}\,{\tiny$\pm$.000} & \underline{0.971}\,{\tiny$\pm$.028} \\
\rowcolor{tabband}SigLIP2-B & \underline{0.183}\,{\tiny$\pm$.050} & \underline{0.060}\,{\tiny$\pm$.019} & 0.020\,{\tiny$\pm$.020} & 0.020\,{\tiny$\pm$.020} & 0.086\,{\tiny$\pm$.047} & \underline{\textbf{0.947}}\,{\tiny$\pm$.018} & \underline{\textbf{1.000}}\,{\tiny$\pm$.000} & \underline{\textbf{1.000}}\,{\tiny$\pm$.000} \\
\addlinespace
Untrained ViT-B & 0.017\,{\tiny$\pm$.017} & 0.007\,{\tiny$\pm$.007} & 0.000\,{\tiny$\pm$.000} & 0.020\,{\tiny$\pm$.020} & 0.000\,{\tiny$\pm$.000} & 0.333\,{\tiny$\pm$.038} & 0.025\,{\tiny$\pm$.025} & 0.086\,{\tiny$\pm$.047} \\
\midrule
\rowcolor{tabband}Qwen3-0.6B & 0.600\,{\tiny$\pm$.063} & 0.687\,{\tiny$\pm$.038} & 0.920\,{\tiny$\pm$.038} & 0.220\,{\tiny$\pm$.059} & 0.943\,{\tiny$\pm$.039} & 0.213\,{\tiny$\pm$.033} & 1.000\,{\tiny$\pm$.000} & 0.771\,{\tiny$\pm$.071} \\
\rowcolor{tabband}Qwen3-4B & 0.983\,{\tiny$\pm$.017} & 0.733\,{\tiny$\pm$.036} & 0.960\,{\tiny$\pm$.028} & 0.220\,{\tiny$\pm$.059} & 0.914\,{\tiny$\pm$.047} & 0.260\,{\tiny$\pm$.036} & 1.000\,{\tiny$\pm$.000} & 0.771\,{\tiny$\pm$.071} \\
\rowcolor{tabband}Qwen3-8B & 0.967\,{\tiny$\pm$.023} & 0.740\,{\tiny$\pm$.036} & 0.960\,{\tiny$\pm$.028} & 0.120\,{\tiny$\pm$.046} & 0.857\,{\tiny$\pm$.059} & 0.240\,{\tiny$\pm$.035} & 1.000\,{\tiny$\pm$.000} & 0.714\,{\tiny$\pm$.076} \\
w2v-BERT~2.0 & 0.000\,{\tiny$\pm$.000} & 0.013\,{\tiny$\pm$.009} & 0.020\,{\tiny$\pm$.020} & 0.000\,{\tiny$\pm$.000} & 0.029\,{\tiny$\pm$.028} & 0.453\,{\tiny$\pm$.041} & 0.225\,{\tiny$\pm$.066} & 0.257\,{\tiny$\pm$.074} \\
\addlinespace
Chance & 0.006\hphantom{\,{\tiny$\pm$.000}} & 0.006\hphantom{\,{\tiny$\pm$.000}} & 0.006\hphantom{\,{\tiny$\pm$.000}} & 0.006\hphantom{\,{\tiny$\pm$.000}} & 0.006\hphantom{\,{\tiny$\pm$.000}} & 0.006\hphantom{\,{\tiny$\pm$.000}} & 0.006\hphantom{\,{\tiny$\pm$.000}} & 0.006\hphantom{\,{\tiny$\pm$.000}} \\
\midrule
\multicolumn{9}{l}{\textbf{(b) Reverse}} \\
\addlinespace[2pt]
\rowcolor{tabband}DINOv3-B & 0.033\,{\tiny$\pm$.023} & 0.000\,{\tiny$\pm$.000} & 0.020\,{\tiny$\pm$.020} & 0.000\,{\tiny$\pm$.000} & 0.029\,{\tiny$\pm$.028} & \underline{0.687}\,{\tiny$\pm$.038} & \underline{0.475}\,{\tiny$\pm$.079} & 0.257\,{\tiny$\pm$.074} \\
\rowcolor{tabband}DINOv3-L & 0.050\,{\tiny$\pm$.028} & 0.013\,{\tiny$\pm$.009} & 0.000\,{\tiny$\pm$.000} & 0.000\,{\tiny$\pm$.000} & 0.057\,{\tiny$\pm$.039} & \underline{0.807}\,{\tiny$\pm$.032} & \underline{\textbf{1.000}}\,{\tiny$\pm$.000} & \underline{0.571}\,{\tiny$\pm$.084} \\
\rowcolor{tabband}DINOv3-H & \underline{0.483}\,{\tiny$\pm$.065} & \underline{0.113}\,{\tiny$\pm$.026} & 0.020\,{\tiny$\pm$.020} & 0.020\,{\tiny$\pm$.020} & \underline{0.114}\,{\tiny$\pm$.054} & \underline{0.653}\,{\tiny$\pm$.039} & \underline{0.950}\,{\tiny$\pm$.034} & \underline{0.800}\,{\tiny$\pm$.068} \\
\rowcolor{tabband}DINOv3-7B & \underline{0.450}\,{\tiny$\pm$.064} & \underline{0.133}\,{\tiny$\pm$.028} & 0.040\,{\tiny$\pm$.028} & 0.020\,{\tiny$\pm$.020} & \underline{0.114}\,{\tiny$\pm$.054} & \underline{0.720}\,{\tiny$\pm$.037} & \underline{\textbf{1.000}}\,{\tiny$\pm$.000} & \underline{0.943}\,{\tiny$\pm$.039} \\
\addlinespace
Web-DINO-1B & 0.000\,{\tiny$\pm$.000} & 0.013\,{\tiny$\pm$.009} & 0.000\,{\tiny$\pm$.000} & 0.000\,{\tiny$\pm$.000} & 0.029\,{\tiny$\pm$.028} & \underline{0.580}\,{\tiny$\pm$.040} & \underline{0.650}\,{\tiny$\pm$.075} & 0.229\,{\tiny$\pm$.071} \\
\addlinespace
\rowcolor{tabband}DINOv2-B & 0.000\,{\tiny$\pm$.000} & 0.007\,{\tiny$\pm$.007} & 0.000\,{\tiny$\pm$.000} & 0.000\,{\tiny$\pm$.000} & 0.029\,{\tiny$\pm$.028} & 0.373\,{\tiny$\pm$.039} & 0.125\,{\tiny$\pm$.052} & 0.000\,{\tiny$\pm$.000} \\
\rowcolor{tabband}DINOv2-L & 0.000\,{\tiny$\pm$.000} & 0.013\,{\tiny$\pm$.009} & 0.040\,{\tiny$\pm$.028} & 0.000\,{\tiny$\pm$.000} & 0.057\,{\tiny$\pm$.039} & 0.407\,{\tiny$\pm$.040} & 0.100\,{\tiny$\pm$.047} & 0.029\,{\tiny$\pm$.028} \\
\rowcolor{tabband}DINOv2-G & 0.000\,{\tiny$\pm$.000} & 0.020\,{\tiny$\pm$.011} & 0.020\,{\tiny$\pm$.020} & 0.000\,{\tiny$\pm$.000} & 0.029\,{\tiny$\pm$.028} & 0.380\,{\tiny$\pm$.040} & 0.125\,{\tiny$\pm$.052} & 0.057\,{\tiny$\pm$.039} \\
\addlinespace
MAE-B & 0.000\,{\tiny$\pm$.000} & 0.013\,{\tiny$\pm$.009} & 0.020\,{\tiny$\pm$.020} & 0.000\,{\tiny$\pm$.000} & 0.000\,{\tiny$\pm$.000} & \underline{0.500}\,{\tiny$\pm$.041} & \underline{0.225}\,{\tiny$\pm$.066} & 0.229\,{\tiny$\pm$.071} \\
MAE-L & 0.000\,{\tiny$\pm$.000} & 0.007\,{\tiny$\pm$.007} & 0.020\,{\tiny$\pm$.020} & 0.000\,{\tiny$\pm$.000} & 0.029\,{\tiny$\pm$.028} & \underline{0.547}\,{\tiny$\pm$.041} & \underline{0.225}\,{\tiny$\pm$.066} & 0.229\,{\tiny$\pm$.071} \\
MAE-H & 0.000\,{\tiny$\pm$.000} & 0.020\,{\tiny$\pm$.011} & 0.020\,{\tiny$\pm$.020} & 0.000\,{\tiny$\pm$.000} & 0.000\,{\tiny$\pm$.000} & \underline{0.573}\,{\tiny$\pm$.040} & \underline{0.275}\,{\tiny$\pm$.071} & \underline{0.343}\,{\tiny$\pm$.080} \\
\addlinespace
\rowcolor{tabband}CLIP-B & \underline{\textbf{0.600}}\,{\tiny$\pm$.063} & \underline{\textbf{0.173}}\,{\tiny$\pm$.031} & \underline{\textbf{0.180}}\,{\tiny$\pm$.054} & \textbf{0.120}\,{\tiny$\pm$.046} & \underline{\textbf{0.143}}\,{\tiny$\pm$.059} & \underline{0.893}\,{\tiny$\pm$.025} & \underline{\textbf{1.000}}\,{\tiny$\pm$.000} & \underline{\textbf{0.971}}\,{\tiny$\pm$.028} \\
\rowcolor{tabband}SigLIP2-B & \underline{0.150}\,{\tiny$\pm$.046} & \underline{0.067}\,{\tiny$\pm$.020} & 0.060\,{\tiny$\pm$.034} & 0.040\,{\tiny$\pm$.028} & \underline{0.114}\,{\tiny$\pm$.054} & \underline{\textbf{0.920}}\,{\tiny$\pm$.022} & \underline{\textbf{1.000}}\,{\tiny$\pm$.000} & \underline{\textbf{0.971}}\,{\tiny$\pm$.028} \\
\addlinespace
Untrained ViT-B & 0.000\,{\tiny$\pm$.000} & 0.000\,{\tiny$\pm$.000} & 0.000\,{\tiny$\pm$.000} & 0.020\,{\tiny$\pm$.020} & 0.000\,{\tiny$\pm$.000} & 0.327\,{\tiny$\pm$.038} & 0.025\,{\tiny$\pm$.025} & 0.086\,{\tiny$\pm$.047} \\
\midrule
\rowcolor{tabband}Qwen3-0.6B & 0.583\,{\tiny$\pm$.064} & 0.680\,{\tiny$\pm$.038} & 0.900\,{\tiny$\pm$.042} & 0.300\,{\tiny$\pm$.065} & 0.971\,{\tiny$\pm$.028} & 0.220\,{\tiny$\pm$.034} & 1.000\,{\tiny$\pm$.000} & 0.857\,{\tiny$\pm$.059} \\
\rowcolor{tabband}Qwen3-4B & 0.983\,{\tiny$\pm$.017} & 0.740\,{\tiny$\pm$.036} & 0.960\,{\tiny$\pm$.028} & 0.260\,{\tiny$\pm$.062} & 0.971\,{\tiny$\pm$.028} & 0.320\,{\tiny$\pm$.038} & 1.000\,{\tiny$\pm$.000} & 0.886\,{\tiny$\pm$.054} \\
\rowcolor{tabband}Qwen3-8B & 0.983\,{\tiny$\pm$.017} & 0.767\,{\tiny$\pm$.035} & 0.960\,{\tiny$\pm$.028} & 0.200\,{\tiny$\pm$.057} & 0.971\,{\tiny$\pm$.028} & 0.273\,{\tiny$\pm$.036} & 1.000\,{\tiny$\pm$.000} & 0.886\,{\tiny$\pm$.054} \\
w2v-BERT~2.0 & 0.000\,{\tiny$\pm$.000} & 0.013\,{\tiny$\pm$.009} & 0.020\,{\tiny$\pm$.020} & 0.020\,{\tiny$\pm$.020} & 0.000\,{\tiny$\pm$.000} & 0.533\,{\tiny$\pm$.041} & 0.250\,{\tiny$\pm$.068} & 0.171\,{\tiny$\pm$.064} \\
\addlinespace
Chance & 0.006\hphantom{\,{\tiny$\pm$.000}} & 0.006\hphantom{\,{\tiny$\pm$.000}} & 0.006\hphantom{\,{\tiny$\pm$.000}} & 0.006\hphantom{\,{\tiny$\pm$.000}} & 0.006\hphantom{\,{\tiny$\pm$.000}} & 0.006\hphantom{\,{\tiny$\pm$.000}} & 0.006\hphantom{\,{\tiny$\pm$.000}} & 0.006\hphantom{\,{\tiny$\pm$.000}} \\
\bottomrule
}%
\setlength{\tabcolsep}{0pt}%
\sbox0{\begin{tabular}{lrrrrrrrr}\tablebody\end{tabular}}%
\ifdim\wd0<\linewidth%
\setlength{\tabcolsep}{\dimexpr(\linewidth-\wd0)/\numexpr2*9\relax\relax}%
\else\setlength{\tabcolsep}{0pt}\fi%
\begin{tabular}{lrrrrrrrr}\tablebody\end{tabular}
\endgroup

%% file: tables/tableB3_items.tex
\definecolor{tabband}{gray}{0.94}
\begingroup
\def\tablebody{%
\toprule
 & \multicolumn{3}{c}{\shortstack{Synonym against\\one-edit word}} & \multicolumn{1}{c}{\shortstack{Synonym\\against\\antonym}} & \multicolumn{1}{c}{\shortstack{Synonym\\against\\base}} & \multicolumn{2}{c}{\shortstack{Top-10 hits\\for synonyms}} \\
\cmidrule(lr){2-4}\cmidrule(lr){5-5}\cmidrule(lr){6-6}\cmidrule(l){7-8}
Model & \shortstack{Count\\of 150} & \shortstack{One-edit\\word} & Synonym & \shortstack{Count\\of 50} & \shortstack{Count\\of 35} & Concrete & Abstract \\
\midrule
\rowcolor{tabband}DINOv3-B & 1 & \underline{2.32}\,{\scriptsize$\pm$.04} & 0.22\,{\scriptsize$\pm$.08} & 31 & 3 & 0 & 0 \\
\rowcolor{tabband}DINOv3-L & 2 & \underline{3.24}\,{\scriptsize$\pm$.06} & 0.34\,{\scriptsize$\pm$.09} & 30 & 3 & 3 & 0 \\
\rowcolor{tabband}DINOv3-H & 14 & \underline{3.01}\,{\scriptsize$\pm$.11} & \underline{0.91}\,{\scriptsize$\pm$.11} & 27 & 2 & 14$^{*}$ & 3 \\
\rowcolor{tabband}DINOv3-7B & 17 & \underline{2.82}\,{\scriptsize$\pm$.10} & \underline{0.88}\,{\scriptsize$\pm$.11} & 23 & 2 & \textbf{17}$^{*}$ & 5 \\
\addlinespace
Web-DINO-1B & 5 & \underline{2.30}\,{\scriptsize$\pm$.06} & 0.31\,{\scriptsize$\pm$.08} & 30 & 8 & 1 & 1 \\
\addlinespace
\rowcolor{tabband}DINOv2-B & 6 & 1.85\,{\scriptsize$\pm$.07} & 0.25\,{\scriptsize$\pm$.08} & 31 & 16 & 1 & 0 \\
\rowcolor{tabband}DINOv2-L & 9 & 2.03\,{\scriptsize$\pm$.07} & 0.27\,{\scriptsize$\pm$.08} & 34$^{*}$ & 19 & 2 & 1 \\
\rowcolor{tabband}DINOv2-G & 12 & 1.60\,{\scriptsize$\pm$.05} & 0.33\,{\scriptsize$\pm$.08} & 31 & 17 & 4 & 1 \\
\addlinespace
MAE-B & 10 & 1.58\,{\scriptsize$\pm$.04} & 0.39\,{\scriptsize$\pm$.08} & 27 & 14 & 1 & 1 \\
MAE-L & 5 & 1.68\,{\scriptsize$\pm$.04} & 0.43\,{\scriptsize$\pm$.08} & 30 & 13 & 1 & 0 \\
MAE-H & 7 & 1.63\,{\scriptsize$\pm$.03} & 0.46\,{\scriptsize$\pm$.08} & 27 & 15 & 3 & 1 \\
\addlinespace
\rowcolor{tabband}CLIP-B & 13 & \underline{5.51}\,{\scriptsize$\pm$.15} & \underline{\textbf{1.75}}\,{\scriptsize$\pm$.14} & 33$^{*}$ & 1 & 15 & \textbf{17} \\
\rowcolor{tabband}SigLIP2-B & 1 & \underline{5.07}\,{\scriptsize$\pm$.09} & \underline{1.01}\,{\scriptsize$\pm$.10} & 29 & 1 & 6 & 8 \\
\addlinespace
Untrained ViT-B & 30 & 1.79\,{\scriptsize$\pm$.12} & 0.33\,{\scriptsize$\pm$.09} & 24 & 15 & 1 & 0 \\
\midrule
\rowcolor{tabband}Qwen3-0.6B & 119 & 2.10\,{\scriptsize$\pm$.12} & 4.58\,{\scriptsize$\pm$.19} & 48 & 22 & 48 & 60 \\
\rowcolor{tabband}Qwen3-4B & 118 & 2.60\,{\scriptsize$\pm$.14} & 5.50\,{\scriptsize$\pm$.22} & 49 & 20 & 55 & 59 \\
\rowcolor{tabband}Qwen3-8B & 124 & 2.45\,{\scriptsize$\pm$.14} & 5.86\,{\scriptsize$\pm$.22} & 50 & 26 & 53 & 65 \\
w2v-BERT~2.0 & 16 & 2.25\,{\scriptsize$\pm$.09} & 0.24\,{\scriptsize$\pm$.09} & 27 & 4 & 2 & 0 \\
\addlinespace
Chance & 75 & 0.00\hphantom{\,{\scriptsize$\pm$.00}} & 0.00\hphantom{\,{\scriptsize$\pm$.00}} & 25 & 17.5 & 0.96 & 0.96 \\
\bottomrule
}%
\setlength{\tabcolsep}{0pt}%
\sbox0{\begin{tabular}{lrrrrrrr}\tablebody\end{tabular}}%
\ifdim\wd0<\linewidth%
\setlength{\tabcolsep}{\dimexpr(\linewidth-\wd0)/\numexpr2*8\relax\relax}%
\else\setlength{\tabcolsep}{0pt}\fi%
\begin{tabular}{lrrrrrrr}\tablebody\end{tabular}
\endgroup

%% file: tables/tableB4_category.tex
\definecolor{tabband}{gray}{0.94}
\begingroup
\def\tablebody{%
\toprule
 & \multicolumn{4}{c}{Neighbor purity} & \multicolumn{4}{c}{Nearest-centroid accuracy} \\
\cmidrule(lr){2-5}\cmidrule(l){6-9}
 & \multicolumn{2}{c}{Nouns} & \multicolumn{2}{c}{Brands} & \multicolumn{2}{c}{Nouns} & \multicolumn{2}{c}{Brands} \\
\cmidrule(lr){2-3}\cmidrule(lr){4-5}\cmidrule(lr){6-7}\cmidrule(l){8-9}
Model & Value & $p$ & Value & $p$ & Value & $p$ & Value & $p$ \\
\midrule
\rowcolor{tabband}DINOv3-B & 0.194\,{\scriptsize$\pm$.012} & 0.0010 & 0.184\,{\scriptsize$\pm$.016} & 0.0345 & 0.214\,{\scriptsize$\pm$.030} & 0.1594 & 0.219\,{\scriptsize$\pm$.042} & 0.2964 \\
\rowcolor{tabband}DINOv3-L & \underline{0.223}\,{\scriptsize$\pm$.014} & 0.0005 & 0.235\,{\scriptsize$\pm$.023} & 0.0005 & \underline{0.297}\,{\scriptsize$\pm$.033} & 0.0005 & \underline{0.281}\,{\scriptsize$\pm$.046} & 0.0165 \\
\rowcolor{tabband}DINOv3-H & \underline{0.269}\,{\scriptsize$\pm$.017} & 0.0005 & \underline{\textbf{0.351}}\,{\scriptsize$\pm$.027} & 0.0005 & \underline{0.365}\,{\scriptsize$\pm$.035} & 0.0005 & \underline{0.427}\,{\scriptsize$\pm$.050} & 0.0005 \\
\rowcolor{tabband}DINOv3-7B & \underline{0.308}\,{\scriptsize$\pm$.020} & 0.0005 & \underline{0.334}\,{\scriptsize$\pm$.031} & 0.0005 & \underline{0.396}\,{\scriptsize$\pm$.035} & 0.0005 & \underline{0.469}\,{\scriptsize$\pm$.051} & 0.0005 \\
\addlinespace
Web-DINO-1B & 0.187\,{\scriptsize$\pm$.012} & 0.0180 & 0.167\,{\scriptsize$\pm$.016} & 0.3873 & 0.198\,{\scriptsize$\pm$.029} & 0.4278 & 0.198\,{\scriptsize$\pm$.041} & 0.4353 \\
\addlinespace
\rowcolor{tabband}DINOv2-B & 0.179\,{\scriptsize$\pm$.012} & 0.0925 & 0.164\,{\scriptsize$\pm$.015} & 0.5182 & 0.245\,{\scriptsize$\pm$.031} & 0.0420 & 0.188\,{\scriptsize$\pm$.040} & 0.5317 \\
\rowcolor{tabband}DINOv2-L & 0.174\,{\scriptsize$\pm$.011} & 0.2884 & 0.167\,{\scriptsize$\pm$.015} & 0.4153 & 0.245\,{\scriptsize$\pm$.031} & 0.0490 & 0.208\,{\scriptsize$\pm$.041} & 0.4123 \\
\rowcolor{tabband}DINOv2-G & 0.181\,{\scriptsize$\pm$.011} & 0.0695 & 0.173\,{\scriptsize$\pm$.014} & 0.1879 & \underline{0.271}\,{\scriptsize$\pm$.032} & 0.0090 & 0.188\,{\scriptsize$\pm$.040} & 0.5447 \\
\addlinespace
MAE-B & 0.182\,{\scriptsize$\pm$.011} & 0.0630 & 0.182\,{\scriptsize$\pm$.015} & 0.0340 & 0.250\,{\scriptsize$\pm$.031} & 0.0325 & 0.167\,{\scriptsize$\pm$.038} & 0.7141 \\
MAE-L & 0.186\,{\scriptsize$\pm$.012} & 0.0230 & 0.188\,{\scriptsize$\pm$.014} & 0.0100 & 0.255\,{\scriptsize$\pm$.031} & 0.0155 & 0.188\,{\scriptsize$\pm$.040} & 0.5042 \\
MAE-H & 0.180\,{\scriptsize$\pm$.012} & 0.0980 & 0.183\,{\scriptsize$\pm$.014} & 0.0340 & 0.229\,{\scriptsize$\pm$.030} & 0.1259 & 0.229\,{\scriptsize$\pm$.043} & 0.0875 \\
\addlinespace
\rowcolor{tabband}CLIP-B & \underline{0.297}\,{\scriptsize$\pm$.019} & 0.0005 & \underline{0.343}\,{\scriptsize$\pm$.027} & 0.0005 & \underline{\textbf{0.474}}\,{\scriptsize$\pm$.036} & 0.0005 & \underline{\textbf{0.500}}\,{\scriptsize$\pm$.051} & 0.0005 \\
\rowcolor{tabband}SigLIP2-B & \underline{\textbf{0.318}}\,{\scriptsize$\pm$.019} & 0.0005 & \underline{0.299}\,{\scriptsize$\pm$.025} & 0.0005 & \underline{0.406}\,{\scriptsize$\pm$.035} & 0.0005 & \underline{0.396}\,{\scriptsize$\pm$.050} & 0.0005 \\
\addlinespace
Untrained ViT-B & 0.183\,{\scriptsize$\pm$.011} & 0.0390 & 0.183\,{\scriptsize$\pm$.016} & 0.0410 & 0.182\,{\scriptsize$\pm$.028} & 0.7336 & 0.146\,{\scriptsize$\pm$.036} & 0.9015 \\
\midrule
\rowcolor{tabband}Qwen3-0.6B & 0.709\,{\scriptsize$\pm$.022} & - & 0.632\,{\scriptsize$\pm$.031} & - & 0.906\,{\scriptsize$\pm$.021} & - & 0.812\,{\scriptsize$\pm$.040} & - \\
\rowcolor{tabband}Qwen3-4B & 0.711\,{\scriptsize$\pm$.021} & - & 0.746\,{\scriptsize$\pm$.028} & - & 0.927\,{\scriptsize$\pm$.019} & - & 0.948\,{\scriptsize$\pm$.023} & - \\
\rowcolor{tabband}Qwen3-8B & 0.711\,{\scriptsize$\pm$.021} & - & 0.743\,{\scriptsize$\pm$.027} & - & 0.922\,{\scriptsize$\pm$.019} & - & 0.938\,{\scriptsize$\pm$.025} & - \\
w2v-BERT~2.0 & 0.201\,{\scriptsize$\pm$.012} & - & 0.178\,{\scriptsize$\pm$.017} & - & 0.266\,{\scriptsize$\pm$.032} & - & 0.135\,{\scriptsize$\pm$.035} & - \\
\addlinespace
Chance & 0.164\hphantom{\,{\scriptsize$\pm$.000}} & - & 0.162\hphantom{\,{\scriptsize$\pm$.000}} & - & 0.167\hphantom{\,{\scriptsize$\pm$.000}} & - & 0.167\hphantom{\,{\scriptsize$\pm$.000}} & - \\
\bottomrule
}%
\setlength{\tabcolsep}{0pt}%
\sbox0{\begin{tabular}{lrrrrrrrr}\tablebody\end{tabular}}%
\ifdim\wd0<\linewidth%
\setlength{\tabcolsep}{\dimexpr(\linewidth-\wd0)/\numexpr2*9\relax\relax}%
\else\setlength{\tabcolsep}{0pt}\fi%
\begin{tabular}{lrrrrrrrr}\tablebody\end{tabular}
\endgroup

%% file: tables/tableB7_lexical_depth.tex
\definecolor{tabband}{gray}{0.94}
\textbf{(a)} Noun neighbor purity.\par\smallskip
\begingroup
\def\tablebody{%
\toprule
 & \multicolumn{9}{c}{Relative depth} \\
\cmidrule(l){2-10}
Model & 0.10 & 0.23 & 0.33 & 0.44 & 0.56 & 0.67 & 0.77 & 0.90 & 1.00 \\
\midrule
\rowcolor{tabband}DINOv3-B & .166\,{\tiny$\pm$.009} & .172\,{\tiny$\pm$.010} & .186\,{\tiny$\pm$.011} & .192\,{\tiny$\pm$.012} & .197\,{\tiny$\pm$.012} & .189\,{\tiny$\pm$.012} & \textbf{.198}\,{\tiny$\pm$.012} & .195\,{\tiny$\pm$.012} & .194\,{\tiny$\pm$.012} \\
\rowcolor{tabband}DINOv3-L & .185\,{\tiny$\pm$.012} & .180\,{\tiny$\pm$.012} & .176\,{\tiny$\pm$.010} & .183\,{\tiny$\pm$.011} & .181\,{\tiny$\pm$.011} & .198\,{\tiny$\pm$.012} & .193\,{\tiny$\pm$.012} & .207\,{\tiny$\pm$.014} & \underline{\textbf{.223}}\,{\tiny$\pm$.014} \\
\rowcolor{tabband}DINOv3-H & .183\,{\tiny$\pm$.011} & .179\,{\tiny$\pm$.011} & .176\,{\tiny$\pm$.011} & .181\,{\tiny$\pm$.011} & .197\,{\tiny$\pm$.012} & .212\,{\tiny$\pm$.013} & .212\,{\tiny$\pm$.014} & \underline{.233}\,{\tiny$\pm$.015} & \underline{\textbf{.269}}\,{\tiny$\pm$.017} \\
\rowcolor{tabband}DINOv3-7B & .180\,{\tiny$\pm$.011} & .187\,{\tiny$\pm$.010} & .181\,{\tiny$\pm$.012} & .199\,{\tiny$\pm$.012} & .209\,{\tiny$\pm$.013} & \underline{.219}\,{\tiny$\pm$.015} & \underline{.233}\,{\tiny$\pm$.015} & \underline{.285}\,{\tiny$\pm$.018} & \underline{\textbf{.308}}\,{\tiny$\pm$.020} \\
\addlinespace
Web-DINO-1B & .178\,{\tiny$\pm$.010} & .179\,{\tiny$\pm$.010} & .176\,{\tiny$\pm$.011} & .189\,{\tiny$\pm$.012} & .197\,{\tiny$\pm$.012} & .198\,{\tiny$\pm$.013} & \textbf{.203}\,{\tiny$\pm$.013} & .200\,{\tiny$\pm$.013} & .187\,{\tiny$\pm$.012} \\
\addlinespace
\rowcolor{tabband}DINOv2-B & .160\,{\tiny$\pm$.010} & .174\,{\tiny$\pm$.010} & .178\,{\tiny$\pm$.011} & .186\,{\tiny$\pm$.011} & .199\,{\tiny$\pm$.013} & .199\,{\tiny$\pm$.013} & \textbf{.199}\,{\tiny$\pm$.012} & .191\,{\tiny$\pm$.012} & .179\,{\tiny$\pm$.012} \\
\rowcolor{tabband}DINOv2-L & .158\,{\tiny$\pm$.010} & .166\,{\tiny$\pm$.011} & .166\,{\tiny$\pm$.010} & .179\,{\tiny$\pm$.011} & .192\,{\tiny$\pm$.011} & .186\,{\tiny$\pm$.011} & \textbf{.199}\,{\tiny$\pm$.013} & .191\,{\tiny$\pm$.012} & .174\,{\tiny$\pm$.011} \\
\rowcolor{tabband}DINOv2-G & .169\,{\tiny$\pm$.010} & .172\,{\tiny$\pm$.010} & .181\,{\tiny$\pm$.011} & \textbf{.187}\,{\tiny$\pm$.012} & \textbf{.187}\,{\tiny$\pm$.012} & .185\,{\tiny$\pm$.012} & .186\,{\tiny$\pm$.012} & .180\,{\tiny$\pm$.011} & .181\,{\tiny$\pm$.011} \\
\addlinespace
MAE-B & .173\,{\tiny$\pm$.011} & .180\,{\tiny$\pm$.011} & \textbf{.188}\,{\tiny$\pm$.010} & .171\,{\tiny$\pm$.011} & .179\,{\tiny$\pm$.010} & .180\,{\tiny$\pm$.010} & .186\,{\tiny$\pm$.011} & .184\,{\tiny$\pm$.011} & .182\,{\tiny$\pm$.011} \\
MAE-L & .185\,{\tiny$\pm$.011} & .180\,{\tiny$\pm$.010} & .183\,{\tiny$\pm$.011} & .184\,{\tiny$\pm$.011} & .188\,{\tiny$\pm$.011} & .192\,{\tiny$\pm$.011} & \textbf{.194}\,{\tiny$\pm$.011} & .188\,{\tiny$\pm$.012} & .186\,{\tiny$\pm$.012} \\
MAE-H & .164\,{\tiny$\pm$.010} & .175\,{\tiny$\pm$.011} & .175\,{\tiny$\pm$.010} & .178\,{\tiny$\pm$.011} & .182\,{\tiny$\pm$.010} & \textbf{.184}\,{\tiny$\pm$.011} & .184\,{\tiny$\pm$.011} & .182\,{\tiny$\pm$.012} & .180\,{\tiny$\pm$.012} \\
\addlinespace
\rowcolor{tabband}CLIP-B & .177\,{\tiny$\pm$.009} & .180\,{\tiny$\pm$.011} & .183\,{\tiny$\pm$.011} & .191\,{\tiny$\pm$.012} & \underline{.217}\,{\tiny$\pm$.014} & \underline{.226}\,{\tiny$\pm$.014} & \underline{.241}\,{\tiny$\pm$.016} & \underline{.289}\,{\tiny$\pm$.019} & \underline{\textbf{.297}}\,{\tiny$\pm$.019} \\
\rowcolor{tabband}SigLIP2-B & .176\,{\tiny$\pm$.010} & .183\,{\tiny$\pm$.011} & .188\,{\tiny$\pm$.012} & .204\,{\tiny$\pm$.012} & \underline{.227}\,{\tiny$\pm$.014} & \underline{.229}\,{\tiny$\pm$.015} & \underline{.250}\,{\tiny$\pm$.018} & \underline{.288}\,{\tiny$\pm$.020} & \underline{\textbf{.318}}\,{\tiny$\pm$.019} \\
\addlinespace
Untrained ViT-B & .171\,{\tiny$\pm$.011} & .175\,{\tiny$\pm$.011} & .176\,{\tiny$\pm$.011} & .176\,{\tiny$\pm$.011} & .179\,{\tiny$\pm$.010} & .180\,{\tiny$\pm$.011} & .179\,{\tiny$\pm$.011} & .182\,{\tiny$\pm$.011} & \textbf{.183}\,{\tiny$\pm$.011} \\
\midrule
Chance & .164\hphantom{\,{\tiny$\pm$.000}} & .164\hphantom{\,{\tiny$\pm$.000}} & .164\hphantom{\,{\tiny$\pm$.000}} & .164\hphantom{\,{\tiny$\pm$.000}} & .164\hphantom{\,{\tiny$\pm$.000}} & .164\hphantom{\,{\tiny$\pm$.000}} & .164\hphantom{\,{\tiny$\pm$.000}} & .164\hphantom{\,{\tiny$\pm$.000}} & .164\hphantom{\,{\tiny$\pm$.000}} \\
\bottomrule
}%
\setlength{\tabcolsep}{0pt}%
\sbox0{\begin{tabular}{lrrrrrrrrr}\tablebody\end{tabular}}%
\ifdim\wd0<\linewidth%
\setlength{\tabcolsep}{\dimexpr(\linewidth-\wd0)/\numexpr2*10\relax\relax}%
\else\setlength{\tabcolsep}{0pt}\fi%
\begin{tabular}{lrrrrrrrrr}\tablebody\end{tabular}
\endgroup
\par\medskip
\textbf{(b)} Capital top-10 accuracy.\par\smallskip
\begingroup
\def\tablebody{%
\toprule
 & \multicolumn{9}{c}{Relative depth} \\
\cmidrule(l){2-10}
Model & 0.10 & 0.23 & 0.33 & 0.44 & 0.56 & 0.67 & 0.77 & 0.90 & 1.00 \\
\midrule
\rowcolor{tabband}DINOv3-B & .000\,{\tiny$\pm$.000} & \textbf{.017}\,{\tiny$\pm$.017} & \textbf{.017}\,{\tiny$\pm$.017} & .000\,{\tiny$\pm$.000} & \textbf{.017}\,{\tiny$\pm$.017} & \textbf{.017}\,{\tiny$\pm$.017} & \textbf{.017}\,{\tiny$\pm$.017} & .000\,{\tiny$\pm$.000} & .000\,{\tiny$\pm$.000} \\
\rowcolor{tabband}DINOv3-L & .017\,{\tiny$\pm$.017} & .000\,{\tiny$\pm$.000} & .000\,{\tiny$\pm$.000} & .033\,{\tiny$\pm$.023} & .017\,{\tiny$\pm$.017} & .017\,{\tiny$\pm$.017} & .017\,{\tiny$\pm$.017} & .033\,{\tiny$\pm$.023} & \underline{\textbf{.117}}\,{\tiny$\pm$.041} \\
\rowcolor{tabband}DINOv3-H & .000\,{\tiny$\pm$.000} & .017\,{\tiny$\pm$.017} & .033\,{\tiny$\pm$.023} & .017\,{\tiny$\pm$.017} & .000\,{\tiny$\pm$.000} & .017\,{\tiny$\pm$.017} & .017\,{\tiny$\pm$.017} & \underline{.167}\,{\tiny$\pm$.048} & \underline{\textbf{.600}}\,{\tiny$\pm$.063} \\
\rowcolor{tabband}DINOv3-7B & .000\,{\tiny$\pm$.000} & .000\,{\tiny$\pm$.000} & .000\,{\tiny$\pm$.000} & .017\,{\tiny$\pm$.017} & .033\,{\tiny$\pm$.023} & .017\,{\tiny$\pm$.017} & .033\,{\tiny$\pm$.023} & \underline{.383}\,{\tiny$\pm$.063} & \underline{\textbf{.600}}\,{\tiny$\pm$.063} \\
\addlinespace
Web-DINO-1B & .017\,{\tiny$\pm$.017} & .017\,{\tiny$\pm$.017} & .017\,{\tiny$\pm$.017} & .017\,{\tiny$\pm$.017} & \textbf{.033}\,{\tiny$\pm$.023} & .017\,{\tiny$\pm$.017} & .000\,{\tiny$\pm$.000} & .000\,{\tiny$\pm$.000} & .000\,{\tiny$\pm$.000} \\
\addlinespace
\rowcolor{tabband}DINOv2-B & .017\,{\tiny$\pm$.017} & .017\,{\tiny$\pm$.017} & \textbf{.033}\,{\tiny$\pm$.023} & \textbf{.033}\,{\tiny$\pm$.023} & .000\,{\tiny$\pm$.000} & .000\,{\tiny$\pm$.000} & .000\,{\tiny$\pm$.000} & .000\,{\tiny$\pm$.000} & .000\,{\tiny$\pm$.000} \\
\rowcolor{tabband}DINOv2-L & .017\,{\tiny$\pm$.017} & .017\,{\tiny$\pm$.017} & .017\,{\tiny$\pm$.017} & \textbf{.050}\,{\tiny$\pm$.028} & .000\,{\tiny$\pm$.000} & .000\,{\tiny$\pm$.000} & .017\,{\tiny$\pm$.017} & .017\,{\tiny$\pm$.017} & .017\,{\tiny$\pm$.017} \\
\rowcolor{tabband}DINOv2-G & .017\,{\tiny$\pm$.017} & .000\,{\tiny$\pm$.000} & \underline{\textbf{.067}}\,{\tiny$\pm$.032} & .033\,{\tiny$\pm$.023} & .017\,{\tiny$\pm$.017} & .000\,{\tiny$\pm$.000} & .000\,{\tiny$\pm$.000} & .000\,{\tiny$\pm$.000} & .000\,{\tiny$\pm$.000} \\
\addlinespace
MAE-B & .017\,{\tiny$\pm$.017} & .000\,{\tiny$\pm$.000} & .000\,{\tiny$\pm$.000} & .000\,{\tiny$\pm$.000} & .000\,{\tiny$\pm$.000} & .000\,{\tiny$\pm$.000} & \textbf{.033}\,{\tiny$\pm$.023} & .000\,{\tiny$\pm$.000} & .000\,{\tiny$\pm$.000} \\
MAE-L & .000\,{\tiny$\pm$.000} & .000\,{\tiny$\pm$.000} & .033\,{\tiny$\pm$.023} & \textbf{.050}\,{\tiny$\pm$.028} & .033\,{\tiny$\pm$.023} & .017\,{\tiny$\pm$.017} & .017\,{\tiny$\pm$.017} & .017\,{\tiny$\pm$.017} & .000\,{\tiny$\pm$.000} \\
MAE-H & .000\,{\tiny$\pm$.000} & .000\,{\tiny$\pm$.000} & .017\,{\tiny$\pm$.017} & \textbf{.033}\,{\tiny$\pm$.023} & \textbf{.033}\,{\tiny$\pm$.023} & \textbf{.033}\,{\tiny$\pm$.023} & .017\,{\tiny$\pm$.017} & .017\,{\tiny$\pm$.017} & .017\,{\tiny$\pm$.017} \\
\addlinespace
\rowcolor{tabband}CLIP-B & .000\,{\tiny$\pm$.000} & .000\,{\tiny$\pm$.000} & .000\,{\tiny$\pm$.000} & .033\,{\tiny$\pm$.023} & .017\,{\tiny$\pm$.017} & .017\,{\tiny$\pm$.017} & .033\,{\tiny$\pm$.023} & .033\,{\tiny$\pm$.023} & \underline{\textbf{.550}}\,{\tiny$\pm$.064} \\
\rowcolor{tabband}SigLIP2-B & .000\,{\tiny$\pm$.000} & .033\,{\tiny$\pm$.023} & .033\,{\tiny$\pm$.023} & .033\,{\tiny$\pm$.023} & .050\,{\tiny$\pm$.028} & .017\,{\tiny$\pm$.017} & .017\,{\tiny$\pm$.017} & .033\,{\tiny$\pm$.023} & \underline{\textbf{.183}}\,{\tiny$\pm$.050} \\
\addlinespace
Untrained ViT-B & \textbf{.033}\,{\tiny$\pm$.023} & .000\,{\tiny$\pm$.000} & .000\,{\tiny$\pm$.000} & .017\,{\tiny$\pm$.017} & .017\,{\tiny$\pm$.017} & \textbf{.033}\,{\tiny$\pm$.023} & \textbf{.033}\,{\tiny$\pm$.023} & .017\,{\tiny$\pm$.017} & .017\,{\tiny$\pm$.017} \\
\midrule
Chance & .006\hphantom{\,{\tiny$\pm$.000}} & .006\hphantom{\,{\tiny$\pm$.000}} & .006\hphantom{\,{\tiny$\pm$.000}} & .006\hphantom{\,{\tiny$\pm$.000}} & .006\hphantom{\,{\tiny$\pm$.000}} & .006\hphantom{\,{\tiny$\pm$.000}} & .006\hphantom{\,{\tiny$\pm$.000}} & .006\hphantom{\,{\tiny$\pm$.000}} & .006\hphantom{\,{\tiny$\pm$.000}} \\
\bottomrule
}%
\setlength{\tabcolsep}{0pt}%
\sbox0{\begin{tabular}{lrrrrrrrrr}\tablebody\end{tabular}}%
\ifdim\wd0<\linewidth%
\setlength{\tabcolsep}{\dimexpr(\linewidth-\wd0)/\numexpr2*10\relax\relax}%
\else\setlength{\tabcolsep}{0pt}\fi%
\begin{tabular}{lrrrrrrrrr}\tablebody\end{tabular}
\endgroup

%% file: appendices/D_word_concept_analyses.tex
\section{Word-Class Alignment}

\subsection{The class names}
\label{app:concepts}

\paragraph{Selection.} A class name is a lowercase word of three to ten letters listed in WordNet for exactly one ImageNet-1k class. We exclude \emph{crane}, for example, because it names both a bird and a machine. We take at most one name per class, preferring frequent and less ambiguous names, and require an estimated frequency of at least one occurrence per million words. We exclude inflected plurals, brands, offensive words, and words whose frequency comes from a proper name. Conventional object names such as \emph{binoculars} are retained. A name must denote its class when read alone, and writing must not dominate the class photographs. We checked these judgments with a language model; they were not independently validated by human raters.

\begin{table}[!b]
\centering
\footnotesize
\setlength{\tabcolsep}{4pt}
\setlength{\belowcaptionskip}{\abovecaptionskip}
\setlength{\abovecaptionskip}{0pt}
\caption{Word-class alignment at the final layer over the 200 class names. All images is the matched cosine similarity. Null mean is the mean over 2,000 random reassignments of classes to words, and $p$ is the one-sided permutation probability. No detected text is the same measure over the 195 names whose classes keep at least 32 images with no detected text, with visual directions rebuilt from those images. Small numbers are standard errors over words. Underlining marks a value in the All images and No detected text columns above the untrained ViT-B by more than twice the standard error of the difference, and bold marks the largest value in those columns.}
\label{tab:concept_alignment}
\input{tables/tableC1_alignment}
\end{table}

\begin{table}[!b]
\centering
\footnotesize
\setlength{\tabcolsep}{2pt}
\setlength{\belowcaptionskip}{\abovecaptionskip}
\setlength{\abovecaptionskip}{0pt}
\caption{Retrieval in both directions at the final layer. Word to class ranks the 1,000 visual directions for each word direction, and class to word ranks the 200 word directions for each named class. Top-10 accuracy is the share of queries whose target is among the ten nearest, and median rank is taken over queries. No detected text gives top-10 accuracy with visual directions rebuilt from images with no detected text, over 195 names and 838 classes. Small numbers are standard errors over queries. Underlining marks an accuracy above the untrained ViT-B by more than twice the standard error of the difference, and bold marks the largest accuracy and the smallest median rank in each column.}
\label{tab:concept_retrieval}
\input{tables/tableC2_retrieval}
\end{table}

\paragraph{The 200 names.} Of 384 candidate names, 224 satisfy the selection rules. We keep the 200 whose classes have the highest share of validation images with no detected text, to reduce the possibility that a rendered word aligns with writing in its class photographs. The names include animals such as \emph{tiger}, devices such as \emph{microphone}, vehicles, buildings, clothing, musical instruments, foods, and landscape words such as \emph{volcano}. Forty-three names also occur in the vocabulary of \secref{sec:rendering} and use the same word images. Their word directions differ because the centering mean in \eqref{eq:tau} is taken over the full vocabulary for lexical comparisons and over the 200 names for word--class comparisons.

\subsection{The class images}
\label{app:concept_images}

The class images come from the ImageNet-1k validation split, as do the 96 backgrounds, which are left out of the class images. \citet{oquab2024dinov2} report removing near-duplicates of benchmark validation sets from the DINOv2 pretraining data, so these images are less likely to be in that data. \citet{simeoni2025dinov3} describe the DINOv3 curation as following that work but do not report the same deduplication. Every class keeps at least 49 images, center-cropped and resized to 256 pixels as the backgrounds were.

\subsection{Alignment and retrieval in full}
\label{app:concept_tables}

\tabref{tab:concept_alignment} gives word-class alignment and \tabref{tab:concept_retrieval} retrieval in both directions, with standard errors and permutation probabilities as in \appref{app:tests}.

\paragraph{Images with no detected text.} Writing in the class images could align a word direction with its class, so we rebuild each class mean from the images in which the CRAFT detector \citep{baek2019craft}, at the settings used for the backgrounds, finds no text. Only classes that keep at least 32 such images enter, and we center the rebuilt class means over these classes to obtain their visual directions. Word directions keep their centering over all 200 names, and the names of the classes left out are dropped.

\paragraph{Background classes.} We exclude the 96 background photographs from the class-image pool, so no photograph is used to estimate both a word direction and a visual direction. A background may still depict the class named by an overlaid word. This occurs for 22 names, each on one background. Recomputing their word directions from the other 95 backgrounds changes the mean matched cosine similarity over these 22 names by less than 0.006 in every model. The largest absolute change is 0.005803, in DINOv3-H. This check measures the effect of including the one matching background.

\paragraph{Class diagnostics.} Two diagnostics describe the class images and exclude no class. Within-class cosine similarity is the mean cosine similarity between the representations of two images of one class, after subtracting the mean representation over all images. In DINOv3-H its median over the 200 named classes is 0.36, and it is lowest for classes such as \emph{bee} and \emph{church}. The second diagnostic is the cosine similarity between a visual direction and the nearest other visual direction. In DINOv3-H it is highest for pairs such as \emph{corn} and \emph{ear}, the class of an ear of corn, and \emph{laptop} and \emph{notebook}. \emph{Ear} and \emph{notebook} have no selected name but remain retrieval candidates, so a word direction for \emph{corn} or \emph{laptop} can rank a near class above its own.

\FloatBarrier
\subsection{Alignment by depth}
\label{app:concept_depth}

\begin{table}[!b]
\centering
\footnotesize
\setlength{\tabcolsep}{3pt}
\setlength{\belowcaptionskip}{\abovecaptionskip}
\setlength{\abovecaptionskip}{0pt}
\caption{Matched cosine similarity (a) and word-to-class top-10 accuracy (b) at nine depths over the 200 class names. Each column is one depth, labeled by the mean over models of the layer index divided by the layer count, and a column's depths differ across models by at most 0.042. The last two columns are one layer apart in the 12-layer models, three in the 24-layer models, and four in the 32- and 40-layer models. Chance is zero in (a) and 0.010 in (b). This table prints no standard errors, so no value is underlined, and bold marks each model's largest value across depths.}
\label{tab:concept_depth}
\input{tables/tableC4_depth}
\end{table}

We repeat word-class alignment and word-to-class retrieval at nine depths (\tabref{tab:concept_depth}). Through relative depth 0.77, matched cosine similarity stays within 0.032 of zero in every model, and the nine models whose final-layer alignment does not exceed its null stay that close at every depth. At every depth before the last, word-to-class top-10 accuracy is at most 0.055 in every model but SigLIP2-B, which at the next-to-last depth already reaches most of its final top-10 accuracy.

\FloatBarrier
\subsection{Two-clue analysis}
\label{app:composition}

\definecolor{ctarget}{HTML}{F5C842}
\definecolor{ctwin}{HTML}{F2B8B5}
\definecolor{csib}{HTML}{B7D3F2}
\definecolor{cother}{HTML}{ECECEC}
\providecommand{\sq}[1]{\textcolor{#1}{\rule[-0.25ex]{1.5ex}{1.5ex}}}
\providecommand{\inkbar}{\textcolor{black!30}{\rule[-0.2ex]{3.6em}{1.3ex}}}

\begin{wraptable}{r}{0.53\textwidth}
\footnotesize
\setlength{\tabcolsep}{4pt}
\renewcommand{\arraystretch}{1.1}
\centering
\setlength{\belowcaptionskip}{\abovecaptionskip}
\setlength{\abovecaptionskip}{0pt}
\caption{Example clue grid. Each row and column supplies one clue, and each cell contains a target word. For \emph{microphone} (yellow), the broad-clue comparison uses \emph{speaker} (pink), and the narrow-clue comparison uses the three other input devices (blue). Gray targets share neither clue and enter only the comparison with all seven alternatives.}
\label{tab:cluegrid}
\begin{tabular}{@{}l cccc@{}}
\toprule
 & \multicolumn{4}{c@{}}{Narrow clue} \\
\cmidrule(l){2-5}
Broad clue & \emph{sound} & \emph{picture} & \emph{text} & \emph{motion} \\
\midrule
\emph{input}  & \cellcolor{ctarget}\textbf{\emph{microphone}} & \cellcolor{csib}\emph{camera} & \cellcolor{csib}\emph{keyboard} & \cellcolor{csib}\emph{sensor} \\
\emph{output} & \cellcolor{ctwin}\emph{speaker} & \cellcolor{cother}\emph{projector} & \cellcolor{cother}\emph{printer} & \cellcolor{cother}\emph{motor} \\
\bottomrule
\end{tabular}
\par\smallskip
{\footnotesize\setlength{\tabcolsep}{0pt}%
\begin{tabular}{@{}l@{~}l@{\qquad}l@{~}l@{}}
\sq{ctarget} & Target     & \sq{csib}   & Narrow comparison \\
\sq{ctwin}   & Broad comparison & \sq{cother} & Neither clue \\
\end{tabular}}
\end{wraptable}

\begin{figure}[!htb]
\centering
\includegraphics[width=\textwidth]{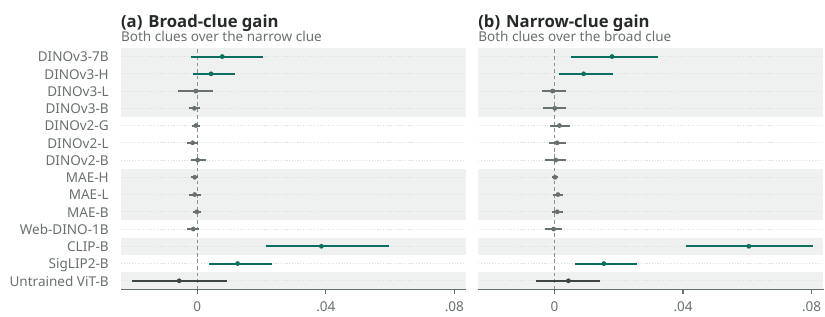}
\caption{\textbf{Contribution of each clue to target-word alignment.} Points show clue gains, with 95\% bootstrap intervals over the twelve grids. The dashed line marks zero gain. Colored points identify the four models whose margin against all seven alternatives has a corrected $p<0.10$; the untrained ViT-B is shown in darker gray. \textbf{(a)} Broad-clue gain: both clues compared with the narrow clue alone. \textbf{(b)} Narrow-clue gain: both clues compared with the broad clue alone.}
\label{fig:composition}
\end{figure}

We test whether two words shown together identify a target more clearly than either word alone. For example, \emph{input} and \emph{sound} together suggest \emph{microphone}. We ask whether each clue improves the pair's alignment with the target relative to competing words. Only CLIP-B shows a contribution from both clues after correction across the thirteen pretrained models (\figref{fig:composition}).

\paragraph{Design.} \tabref{tab:cluegrid} illustrates how we choose the clues and targets. Each grid combines two broad clues with four narrow clues, giving eight targets. In this example, the broad clue specifies whether a device takes a signal in or gives it out, and the narrow clue specifies the signal type. Neither clue identifies a target on its own: \emph{sound} could mean \emph{microphone} or \emph{speaker}, while \emph{input} could mean any of the four devices in the first row. The analysis includes 96 items from twelve grids. Six pilot grids used during development are evaluated separately and excluded from the reported results.

\paragraph{Renderings.} \tabref{tab:renderings} shows the four versions of each clue pair: both clues intact, only the broad clue intact, only the narrow clue intact, and neither intact. We replace each absent clue with a phase-randomized version of its letters, following the control procedure in \appref{app:data}. We render all four versions on the same 96 backgrounds and render the target word separately with its own control. For each condition, we subtract the control representation and average the differences over backgrounds. With $f$ denoting the image encoder, we abbreviate these averages as
\begin{equation}
\begin{aligned}
\hat{\tau}_{c} &= f(c \text{ intact}) - f(\text{neither intact}),
&&\quad c \in \{\text{both},\ \text{broad},\ \text{narrow}\}, \\
\hat{\tau}_{\mathrm{target}} &= f(\text{target}) - f(\text{target control}).
\end{aligned}
\label{eq:composition}
\end{equation}
We compare these vectors only with the eight target words in the same grid and do not center them over a vocabulary.

\begin{tablock}
\footnotesize
\setlength{\tabcolsep}{4pt}
\renewcommand{\arraystretch}{1.1}
\centering
\setlength{\belowcaptionskip}{\abovecaptionskip}
\setlength{\abovecaptionskip}{0pt}
\captionof{table}{Renderings for the clue pair \emph{input} and \emph{sound} and its target, \emph{microphone}. Gray bars indicate phase-randomized letters. Difference vectors are defined in \eqref{eq:composition}.}
\label{tab:renderings}
\begin{tabular}{@{}l cc >{\raggedright\arraybackslash}p{3.1in}@{}}
\toprule
Rendering & Broad clue & Narrow clue & Difference vector \\
\midrule
Both intact    & \emph{input} & \emph{sound} & $\hat{\tau}_{\mathrm{both}}$, used in both clue comparisons \\
Broad intact   & \emph{input} & \inkbar      & $\hat{\tau}_{\mathrm{broad}}$, baseline for the narrow-clue gain \\
Narrow intact  & \inkbar      & \emph{sound} & $\hat{\tau}_{\mathrm{narrow}}$, baseline for the broad-clue gain \\
Neither intact & \inkbar      & \inkbar      & The control subtracted from the three difference vectors above \\
\midrule
Target alone   & \multicolumn{2}{c}{\emph{microphone}} & $\hat{\tau}_{\mathrm{target}}$, against its own control image \\
\bottomrule
\end{tabular}
\end{tablock}

\paragraph{Clue gains.} We measure each clue's contribution by comparing the target with words that share the other clue. For \emph{microphone}, the broad clue \emph{input} must distinguish it from \emph{speaker}, which also matches \emph{sound}. The narrow clue \emph{sound} must distinguish it from \emph{camera}, \emph{keyboard}, and \emph{sensor}, which all match \emph{input}. We call these the broad and narrow comparisons, respectively.

For each clue condition, the margin is the cosine similarity of its difference vector to the target vector minus the mean similarity to the competing target vectors. The clue gain is the increase in this margin when the missing clue is added. Thus, the broad-clue gain compares $\hat{\tau}_{\mathrm{both}}$ with $\hat{\tau}_{\mathrm{narrow}}$, and the narrow-clue gain compares $\hat{\tau}_{\mathrm{both}}$ with $\hat{\tau}_{\mathrm{broad}}$. We report the smaller gain and apply Holm correction to the larger of the two $p$-values across the thirteen pretrained models. This requires evidence that both clues contribute. A single margin against all seven alternatives cannot establish this, because the narrow clue alone distinguishes the target from six of them.

\paragraph{Results.} \figref{fig:composition} shows the gain from adding each clue. CLIP-B is the only model with significant gains for both clues after correction. SigLIP2-B has bootstrap intervals above zero for both gains, but does not pass the corrected test. DINOv3-H and DINOv3-7B have intervals above zero only for the narrow-clue gain, and the reported tests do not establish that these gains exceed the untrained ViT-B's. For these two models, the test detects a contribution from the narrow clue but no contribution from the broad clue that distinguishes targets sharing the narrow clue.

We also test whether the pair contributes anything beyond the sum of the two single-clue vectors. We subtract that sum from the pair vector and measure the residual's margin against all seven competing targets. No model has a residual margin significantly different from zero after correction. CLIP-B and MAE-H have negative residual margins with bootstrap intervals below zero, but these do not pass the corrected test either. The absence of a significant residual margin does not establish that the pair vector equals the sum of the single-clue vectors.

\paragraph{Limits.} These comparisons concern alignment with a rendered target word; they do not test alignment with images of the object it names. Word co-occurrence during training could also account for the gains. The items were selected by the authors, and the small number of independent grids limits generalization. Visual differences beyond letter identity may also affect the comparison: they occur at both clue positions in the pair image, but at only one in each single-clue image.

%% file: tables/tableC1_alignment.tex
\definecolor{tabband}{gray}{0.94}
\begingroup
\def\tablebody{%
\toprule
 & \multicolumn{3}{c}{Matched cosine similarity} & \\
\cmidrule(lr){2-4}
Model & All images & Null mean & $p$ & \shortstack{No detected\\text} \\
\midrule
\rowcolor{tabband}DINOv3-B & 0.002\,{\scriptsize$\pm$.003} & -0.0001 & 0.237 & 0.002\,{\scriptsize$\pm$.003} \\
\rowcolor{tabband}DINOv3-L & \underline{0.025}\,{\scriptsize$\pm$.004} & -0.0003 & $<$0.001 & 0.023\,{\scriptsize$\pm$.004} \\
\rowcolor{tabband}DINOv3-H & \underline{0.234}\,{\scriptsize$\pm$.014} & -0.0004 & $<$0.001 & \underline{0.230}\,{\scriptsize$\pm$.015} \\
\rowcolor{tabband}DINOv3-7B & \underline{0.120}\,{\scriptsize$\pm$.012} & -0.0045 & $<$0.001 & \underline{0.116}\,{\scriptsize$\pm$.013} \\
\addlinespace
Web-DINO-1B & 0.001\,{\scriptsize$\pm$.002} & -0.0001 & 0.306 & 0.001\,{\scriptsize$\pm$.002} \\
\addlinespace
\rowcolor{tabband}DINOv2-B & -0.002\,{\scriptsize$\pm$.003} & 0.0000 & 0.747 & -0.002\,{\scriptsize$\pm$.003} \\
\rowcolor{tabband}DINOv2-L & -0.001\,{\scriptsize$\pm$.002} & 0.0000 & 0.667 & -0.001\,{\scriptsize$\pm$.002} \\
\rowcolor{tabband}DINOv2-G & 0.001\,{\scriptsize$\pm$.002} & -0.0001 & 0.300 & 0.001\,{\scriptsize$\pm$.002} \\
\addlinespace
MAE-B & -0.011\,{\scriptsize$\pm$.006} & -0.0002 & 0.958 & -0.011\,{\scriptsize$\pm$.006} \\
MAE-L & 0.003\,{\scriptsize$\pm$.004} & 0.0001 & 0.254 & 0.004\,{\scriptsize$\pm$.004} \\
MAE-H & 0.002\,{\scriptsize$\pm$.004} & 0.0004 & 0.317 & 0.003\,{\scriptsize$\pm$.004} \\
\addlinespace
\rowcolor{tabband}CLIP-B & \underline{\textbf{0.237}}\,{\scriptsize$\pm$.010} & 0.0001 & $<$0.001 & \underline{\textbf{0.233}}\,{\scriptsize$\pm$.010} \\
\rowcolor{tabband}SigLIP2-B & \underline{0.184}\,{\scriptsize$\pm$.009} & 0.0000 & $<$0.001 & \underline{0.182}\,{\scriptsize$\pm$.009} \\
\addlinespace
Untrained ViT-B & 0.005\,{\scriptsize$\pm$.008} & 0.0030 & 0.419 & 0.006\,{\scriptsize$\pm$.008} \\
\midrule
Chance & 0.000\hphantom{\,{\scriptsize$\pm$.000}} & & & 0.000\hphantom{\,{\scriptsize$\pm$.000}} \\
\bottomrule
}%
\setlength{\tabcolsep}{0pt}%
\sbox0{\begin{tabular}{lrrrr}\tablebody\end{tabular}}%
\ifdim\wd0<\linewidth%
\setlength{\tabcolsep}{\dimexpr(\linewidth-\wd0)/\numexpr2*5\relax\relax}%
\else\setlength{\tabcolsep}{0pt}\fi%
\begin{tabular}{lrrrr}\tablebody\end{tabular}
\endgroup

%% file: tables/tableC2_retrieval.tex
\definecolor{tabband}{gray}{0.94}
\begingroup
\def\tablebody{%
\toprule
 & \multicolumn{3}{c}{Word-to-class retrieval, 1,000 classes} & \multicolumn{3}{c}{Class-to-word retrieval, 200 class names} \\
\cmidrule(lr){2-4}\cmidrule(l){5-7}
Model & \shortstack{Top-10\\accuracy} & \shortstack{Median\\rank} & \shortstack{No detected\\text} & \shortstack{Top-10\\accuracy} & \shortstack{Median\\rank} & \shortstack{No detected\\text} \\
\midrule
\rowcolor{tabband}DINOv3-B & 0.010\,{\scriptsize$\pm$.007} & 486 & 0.010\,{\scriptsize$\pm$.007} & 0.060\,{\scriptsize$\pm$.017} & 94.5 & 0.067\,{\scriptsize$\pm$.018} \\
\rowcolor{tabband}DINOv3-L & \underline{0.120}\,{\scriptsize$\pm$.023} & 335 & \underline{0.103}\,{\scriptsize$\pm$.022} & \underline{0.180}\,{\scriptsize$\pm$.027} & 68 & \underline{0.179}\,{\scriptsize$\pm$.027} \\
\rowcolor{tabband}DINOv3-H & \underline{0.645}\,{\scriptsize$\pm$.034} & \textbf{1} & \underline{0.651}\,{\scriptsize$\pm$.034} & \underline{0.680}\,{\scriptsize$\pm$.033} & \textbf{1} & \underline{0.677}\,{\scriptsize$\pm$.033} \\
\rowcolor{tabband}DINOv3-7B & \underline{0.325}\,{\scriptsize$\pm$.033} & 154 & \underline{0.328}\,{\scriptsize$\pm$.034} & \underline{0.405}\,{\scriptsize$\pm$.035} & 32.5 & \underline{0.390}\,{\scriptsize$\pm$.035} \\
\addlinespace
Web-DINO-1B & 0.020\,{\scriptsize$\pm$.010} & 477 & 0.015\,{\scriptsize$\pm$.009} & 0.045\,{\scriptsize$\pm$.015} & 95 & 0.041\,{\scriptsize$\pm$.014} \\
\addlinespace
\rowcolor{tabband}DINOv2-B & 0.010\,{\scriptsize$\pm$.007} & 527.5 & 0.010\,{\scriptsize$\pm$.007} & 0.060\,{\scriptsize$\pm$.017} & 103.5 & 0.062\,{\scriptsize$\pm$.017} \\
\rowcolor{tabband}DINOv2-L & 0.000\,{\scriptsize$\pm$.000} & 485 & 0.005\,{\scriptsize$\pm$.005} & 0.045\,{\scriptsize$\pm$.015} & 99.5 & 0.056\,{\scriptsize$\pm$.017} \\
\rowcolor{tabband}DINOv2-G & 0.005\,{\scriptsize$\pm$.005} & 463 & 0.005\,{\scriptsize$\pm$.005} & 0.045\,{\scriptsize$\pm$.015} & 94 & 0.062\,{\scriptsize$\pm$.017} \\
\addlinespace
MAE-B & 0.005\,{\scriptsize$\pm$.005} & 559 & 0.005\,{\scriptsize$\pm$.005} & 0.020\,{\scriptsize$\pm$.010} & 113 & 0.026\,{\scriptsize$\pm$.011} \\
MAE-L & 0.015\,{\scriptsize$\pm$.009} & 491 & 0.015\,{\scriptsize$\pm$.009} & 0.045\,{\scriptsize$\pm$.015} & 98.5 & 0.056\,{\scriptsize$\pm$.017} \\
MAE-H & 0.005\,{\scriptsize$\pm$.005} & 461.5 & 0.015\,{\scriptsize$\pm$.009} & 0.050\,{\scriptsize$\pm$.015} & 96.5 & 0.056\,{\scriptsize$\pm$.017} \\
\addlinespace
\rowcolor{tabband}CLIP-B & \underline{\textbf{0.855}}\,{\scriptsize$\pm$.025} & \textbf{1} & \underline{\textbf{0.851}}\,{\scriptsize$\pm$.025} & \underline{\textbf{0.845}}\,{\scriptsize$\pm$.026} & \textbf{1} & \underline{\textbf{0.831}}\,{\scriptsize$\pm$.027} \\
\rowcolor{tabband}SigLIP2-B & \underline{0.630}\,{\scriptsize$\pm$.034} & 2 & \underline{0.636}\,{\scriptsize$\pm$.034} & \underline{0.735}\,{\scriptsize$\pm$.031} & 2 & \underline{0.718}\,{\scriptsize$\pm$.032} \\
\addlinespace
Untrained ViT-B & 0.005\,{\scriptsize$\pm$.005} & 511.5 & 0.005\,{\scriptsize$\pm$.005} & 0.045\,{\scriptsize$\pm$.015} & 99 & 0.046\,{\scriptsize$\pm$.015} \\
\midrule
Chance & 0.010\hphantom{\,{\scriptsize$\pm$.000}} &  & 0.012\hphantom{\,{\scriptsize$\pm$.000}} & 0.050\hphantom{\,{\scriptsize$\pm$.000}} &  & 0.051\hphantom{\,{\scriptsize$\pm$.000}} \\
\bottomrule
}%
\setlength{\tabcolsep}{0pt}%
\sbox0{\begin{tabular}{lrrrrrr}\tablebody\end{tabular}}%
\ifdim\wd0<\linewidth%
\setlength{\tabcolsep}{\dimexpr(\linewidth-\wd0)/\numexpr2*7\relax\relax}%
\else\setlength{\tabcolsep}{0pt}\fi%
\begin{tabular}{lrrrrrr}\tablebody\end{tabular}
\endgroup

%% file: tables/tableC4_depth.tex
\definecolor{tabband}{gray}{0.94}
\textbf{(a)} Matched cosine similarity.\par\smallskip
\begingroup
\def\tablebody{%
\toprule
 & \multicolumn{9}{c}{Relative depth} \\
\cmidrule(l){2-10}
Model & 0.10 & 0.23 & 0.33 & 0.44 & 0.56 & 0.67 & 0.77 & 0.90 & 1.00 \\
\midrule
\rowcolor{tabband}DINOv3-B & -0.019 & 0.011 & \textbf{0.022} & 0.018 & -0.003 & 0.012 & 0.007 & -0.002 & 0.002 \\
\rowcolor{tabband}DINOv3-L & 0.008 & 0.007 & 0.006 & -0.002 & 0.003 & 0.002 & 0.012 & 0.013 & \textbf{0.025} \\
\rowcolor{tabband}DINOv3-H & 0.016 & 0.029 & -0.005 & -0.003 & 0.000 & 0.005 & 0.013 & 0.041 & \textbf{0.234} \\
\rowcolor{tabband}DINOv3-7B & 0.007 & 0.027 & 0.022 & 0.006 & 0.002 & 0.003 & 0.003 & -0.006 & \textbf{0.120} \\
\addlinespace
Web-DINO-1B & 0.014 & \textbf{0.023} & 0.020 & 0.007 & 0.005 & 0.007 & 0.003 & 0.003 & 0.001 \\
\addlinespace
\rowcolor{tabband}DINOv2-B & -0.003 & \textbf{0.011} & 0.010 & 0.005 & -0.009 & -0.010 & -0.009 & -0.002 & -0.002 \\
\rowcolor{tabband}DINOv2-L & -0.004 & 0.014 & \textbf{0.016} & 0.011 & 0.001 & -0.000 & 0.000 & -0.004 & -0.001 \\
\rowcolor{tabband}DINOv2-G & 0.005 & \textbf{0.026} & 0.021 & 0.005 & 0.007 & -0.002 & -0.001 & 0.001 & 0.001 \\
\addlinespace
MAE-B & -0.011 & \textbf{0.031} & 0.022 & 0.028 & 0.019 & 0.017 & 0.005 & -0.005 & -0.011 \\
MAE-L & \textbf{0.026} & 0.021 & 0.018 & 0.013 & 0.008 & 0.004 & -0.000 & -0.001 & 0.003 \\
MAE-H & 0.008 & 0.017 & 0.014 & \textbf{0.019} & 0.008 & 0.003 & 0.003 & 0.003 & 0.002 \\
\addlinespace
\rowcolor{tabband}CLIP-B & 0.022 & 0.030 & 0.022 & 0.014 & 0.016 & 0.007 & 0.003 & 0.036 & \textbf{0.237} \\
\rowcolor{tabband}SigLIP2-B & 0.032 & 0.020 & 0.013 & 0.011 & -0.001 & -0.002 & 0.008 & 0.114 & \textbf{0.184} \\
\addlinespace
Untrained ViT-B & 0.002 & 0.002 & 0.000 & \textbf{0.007} & 0.006 & 0.004 & 0.005 & 0.006 & 0.005 \\
\bottomrule
}%
\setlength{\tabcolsep}{0pt}%
\sbox0{\begin{tabular}{lrrrrrrrrr}\tablebody\end{tabular}}%
\ifdim\wd0<\linewidth%
\setlength{\tabcolsep}{\dimexpr(\linewidth-\wd0)/\numexpr2*10\relax\relax}%
\else\setlength{\tabcolsep}{0pt}\fi%
\begin{tabular}{lrrrrrrrrr}\tablebody\end{tabular}
\endgroup
\par\medskip
\textbf{(b)} Word-to-class top-10 accuracy.\par\smallskip
\begingroup
\def\tablebody{%
\toprule
 & \multicolumn{9}{c}{Relative depth} \\
\cmidrule(l){2-10}
Model & 0.10 & 0.23 & 0.33 & 0.44 & 0.56 & 0.67 & 0.77 & 0.90 & 1.00 \\
\midrule
\rowcolor{tabband}DINOv3-B & 0.005 & \textbf{0.020} & 0.005 & 0.015 & 0.005 & 0.015 & \textbf{0.020} & 0.010 & 0.010 \\
\rowcolor{tabband}DINOv3-L & 0.000 & 0.000 & 0.005 & 0.010 & 0.005 & 0.020 & 0.025 & 0.030 & \textbf{0.120} \\
\rowcolor{tabband}DINOv3-H & 0.015 & 0.020 & 0.010 & 0.010 & 0.005 & 0.015 & 0.010 & 0.050 & \textbf{0.645} \\
\rowcolor{tabband}DINOv3-7B & 0.010 & 0.000 & 0.005 & 0.025 & 0.010 & 0.005 & 0.010 & 0.015 & \textbf{0.325} \\
\addlinespace
Web-DINO-1B & 0.000 & 0.010 & 0.000 & 0.005 & 0.005 & \textbf{0.025} & 0.020 & 0.010 & 0.020 \\
\addlinespace
\rowcolor{tabband}DINOv2-B & 0.005 & 0.005 & 0.005 & 0.010 & \textbf{0.020} & \textbf{0.020} & 0.015 & 0.005 & 0.010 \\
\rowcolor{tabband}DINOv2-L & \textbf{0.010} & \textbf{0.010} & 0.000 & 0.005 & 0.005 & 0.000 & 0.000 & 0.000 & 0.000 \\
\rowcolor{tabband}DINOv2-G & 0.005 & 0.005 & \textbf{0.015} & 0.005 & 0.005 & 0.000 & 0.000 & 0.005 & 0.005 \\
\addlinespace
MAE-B & 0.010 & 0.010 & 0.000 & 0.010 & \textbf{0.020} & 0.015 & 0.010 & 0.010 & 0.005 \\
MAE-L & \textbf{0.020} & 0.010 & 0.005 & \textbf{0.020} & 0.015 & 0.005 & 0.015 & 0.010 & 0.015 \\
MAE-H & 0.000 & 0.010 & \textbf{0.015} & 0.010 & 0.000 & \textbf{0.015} & 0.010 & 0.005 & 0.005 \\
\addlinespace
\rowcolor{tabband}CLIP-B & 0.005 & 0.000 & 0.015 & 0.005 & 0.005 & 0.005 & 0.015 & 0.055 & \textbf{0.855} \\
\rowcolor{tabband}SigLIP2-B & 0.020 & 0.010 & 0.000 & 0.005 & 0.005 & 0.005 & 0.020 & 0.465 & \textbf{0.630} \\
\addlinespace
Untrained ViT-B & 0.010 & \textbf{0.015} & 0.000 & 0.000 & 0.005 & 0.005 & 0.005 & 0.005 & 0.005 \\
\midrule
Chance & 0.010 & 0.010 & 0.010 & 0.010 & 0.010 & 0.010 & 0.010 & 0.010 & 0.010 \\
\bottomrule
}%
\setlength{\tabcolsep}{0pt}%
\sbox0{\begin{tabular}{lrrrrrrrrr}\tablebody\end{tabular}}%
\ifdim\wd0<\linewidth%
\setlength{\tabcolsep}{\dimexpr(\linewidth-\wd0)/\numexpr2*10\relax\relax}%
\else\setlength{\tabcolsep}{0pt}\fi%
\begin{tabular}{lrrrrrrrrr}\tablebody\end{tabular}
\endgroup

%% file: appendices/E_finetuning_and_reproducibility.tex
\section{Models, Encoders, and Finetuning}
\label{app:finetuning}

\subsection{Models and representations}
\label{app:features}

\paragraph{Representation.} \tabref{tab:feature_models} lists the fourteen models and the four encoders of \appref{app:encoders}. The representation $f$ of \secref{sec:word_vectors} is the hidden state after the final transformer block, before the model's final layer normalization. It is the class token in every model but SigLIP2-B, which has no class token and uses the mean of the patch tokens, and neither CLIP-B's visual projection nor SigLIP2-B's attention pooling head is applied.

\begin{tablock}
\centering
\footnotesize
\setlength{\tabcolsep}{4pt}
\setlength{\belowcaptionskip}{\abovecaptionskip}
\setlength{\abovecaptionskip}{0pt}
\captionof{table}{The fourteen models and, under the rule, the three text
encoders and the speech encoder, with the checkpoint identifier, layer count,
and width of each. The untrained ViT-B takes only its architecture from the
listed configuration, and its weights are randomly initialized. A speech vector
concatenates four relative-position bins of the listed width, so it has 4,096
dimensions.}
\label{tab:feature_models}
\begin{tabular}{lp{3.45in}rr}
\toprule
Model & Checkpoint or configuration identifier & Layers & Width \\
\midrule
\rowcolor{tabband}DINOv3-B & \path{facebook/dinov3-vitb16-pretrain-lvd1689m} & 12 & 768 \\
\rowcolor{tabband}DINOv3-L & \path{facebook/dinov3-vitl16-pretrain-lvd1689m} & 24 & 1024 \\
\rowcolor{tabband}DINOv3-H & \path{facebook/dinov3-vith16plus-pretrain-lvd1689m} & 32 & 1280 \\
\rowcolor{tabband}DINOv3-7B & \path{facebook/dinov3-vit7b16-pretrain-lvd1689m} & 40 & 4096 \\
Web-DINO-1B & \path{facebook/webssl-dino1b-full2b-224} & 40 & 1536 \\
\rowcolor{tabband}DINOv2-B & \path{facebook/dinov2-base} & 12 & 768 \\
\rowcolor{tabband}DINOv2-L & \path{facebook/dinov2-large} & 24 & 1024 \\
\rowcolor{tabband}DINOv2-G & \path{facebook/dinov2-giant} & 40 & 1536 \\
MAE-B & \path{facebook/vit-mae-base} & 12 & 768 \\
MAE-L & \path{facebook/vit-mae-large} & 24 & 1024 \\
MAE-H & \path{facebook/vit-mae-huge} & 32 & 1280 \\
\rowcolor{tabband}CLIP-B & \path{openai/clip-vit-base-patch16} & 12 & 768 \\
\rowcolor{tabband}SigLIP2-B & \path{google/siglip2-base-patch16-224} & 12 & 768 \\
Untrained ViT-B & \path{google/vit-base-patch16-224} (configuration only) & 12 & 768 \\
\midrule
\rowcolor{tabband}Qwen3-0.6B & \path{Qwen/Qwen3-Embedding-0.6B} & 28 & 1024 \\
\rowcolor{tabband}Qwen3-4B & \path{Qwen/Qwen3-Embedding-4B} & 36 & 2560 \\
\rowcolor{tabband}Qwen3-8B & \path{Qwen/Qwen3-Embedding-8B} & 36 & 4096 \\
w2v-BERT~2.0 & \path{facebook/w2v-bert-2.0} & 24 & 1024 \\
\bottomrule
\end{tabular}
\end{tablock}

\paragraph{Input processing.} Every image is $256\times256$ pixels, and we give each model the preprocessing published with its checkpoint. For DINOv2 the image is center-cropped to $224\times224$, which keeps the text at its rendered size and trims the edge of the background. For the other models it is resized to $224\times224$, with bicubic interpolation for Web-DINO-1B and CLIP-B and bilinear interpolation for the rest. Color channels are normalized with the means and standard deviations published with each checkpoint.

\paragraph{Averaging and precision.} Representations are not scaled to unit length before paired subtraction or class averaging. Every model runs in float32 except DINOv3-7B, which runs in bfloat16. MAE runs with a mask ratio of zero, so every patch is kept. Depth profiles read the same token at nine depths, from about a tenth of the blocks to the final block (\tabref{tab:concept_depth}). The untrained ViT-B has the ViT-B architecture of \tabref{tab:feature_models}, with 12 attention heads and $16\times16$ patches, one random initialization, and the preprocessing of the ViT-B checkpoint.

\subsection{Text and speech vectors}
\label{app:encoders}

\paragraph{Carriers.} A text or speech encoder never sees a rendered image, so its vector for a word, name, or pseudoword is built by placing it inside a fixed set of carrier sentences, as a word direction is built on a fixed set of backgrounds. The set is the same across the vocabulary, because a carrier chosen per word, such as a sentence in which the word occurs, would bring corpus co-occurrence into the measurement. The carriers vary in syntactic type, because an average keeps whatever its carriers share, and carriers of one type would move every average toward that type.

\paragraph{Text.} Each of the 1,561 words, names, and pseudowords is placed in 96 carriers, the 24 templates of \tabref{tab:carriers} in four variants each. The variants differ in punctuation, tense, or a function word, as in ``The word is \emph{w}.'' and ``The word was \emph{w}.''. The templates supply the variation and the variants are replicates, as the typography settings and their four backgrounds are for the word images. Every target position is quoted, named, or otherwise permissive, so no carrier is ungrammatical for an adjective, a verb, a brand, a capital, or a pseudoword. A carrier's vector is the frozen encoder's state at the last token, the pooling that Qwen3-Embedding-0.6B, -4B, and -8B were trained with. We add no instruction prefix, since the prefix is meant only for queries in asymmetric retrieval, and this comparison is symmetric.

\begin{tablock}
\centering
\footnotesize
\setlength{\tabcolsep}{5pt}
\setlength{\belowcaptionskip}{\abovecaptionskip}
\setlength{\abovecaptionskip}{0pt}
\captionof{table}{The 24 carrier templates for the text encoders, shown in
their first variant with \emph{w} standing for the word, name, or pseudoword. Each template has
four variants, for 96 carriers per word.}
\label{tab:carriers}
\begin{tabular}{@{}ll@{\hspace{14pt}}ll@{}}
\toprule
\# & Template & \# & Template \\
\midrule
1 & \emph{w} & 13 & They wrote \emph{w}. \\
2 & ``\emph{w}'' & 14 & Someone said \emph{w}. \\
3 & the word ``\emph{w}'' & 15 & She read \emph{w} out. \\
4 & The word is \emph{w}. & 16 & He typed \emph{w} quickly. \\
5 & Read \emph{w} aloud. & 17 & I heard \emph{w} again. \\
6 & Spell \emph{w}. & 18 & \emph{w}, and then \emph{w}. \\
7 & \emph{w} is spelled out. & 19 & First \emph{w}. Then the rest. \\
8 & It says \emph{w} there. & 20 & \emph{w}, \emph{w}, \emph{w}. \\
9 & \emph{w} is what it says. & 21 & Next up: \emph{w}. \\
10 & This one is \emph{w}. & 22 & Here it is, \emph{w}. \\
11 & That would be \emph{w}. & 23 & A sign showing \emph{w}. \\
12 & The answer is \emph{w}. & 24 & Written on it: \emph{w}. \\
\bottomrule
\end{tabular}
\end{tablock}

\paragraph{Speech.} Each word, name, or pseudoword is spoken in 96 utterances by a VITS synthesizer trained on VCTK, in eight voices at three speaking rates (0.9, 1.0, and 1.1) and four carrier sentences that place the word first, in the middle, or last: ``\emph{w} was written there.'', ``They wrote \emph{w} on it.'', ``Someone said \emph{w} out loud.'', and ``The label reads \emph{w}.''. Synthesis adds no sampling noise, so an utterance is fixed by its voice, rate, and carrier. The frozen encoder is w2v-BERT~2.0 \citep{seamlessm4t2023}, and only the frames of the word are kept, located by the synthesizer's phoneme durations. These frames are pooled into four bins of equal relative position and concatenated, so the vector keeps where in the word each sound falls, which a plain mean over frames would discard. For text and speech alike, the 96 vectors for each word are averaged and the mean is centered over the full vocabulary of 1,561 words, names, and pseudowords.

\subsection{Contrastive finetuning}
\label{app:finetune_recipe}

\begin{figure}[t]
\centering
\includegraphics[width=\textwidth]{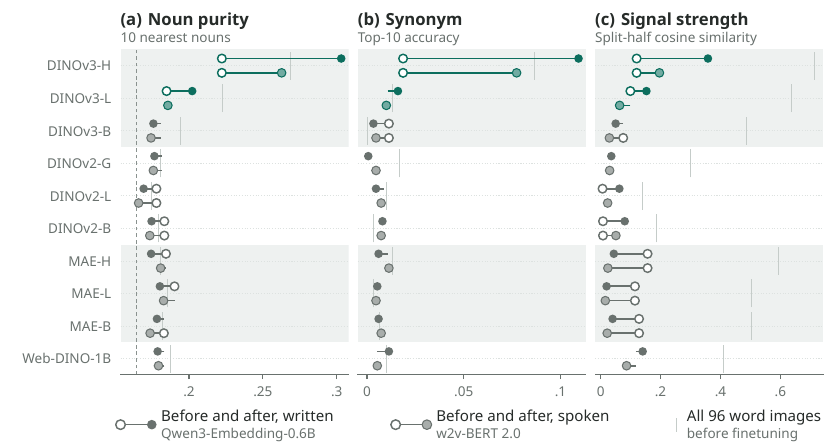}
\caption{\textbf{Finetuning raises noun purity and synonym retrieval in DINOv3-H.} Open circles show scores before finetuning; filled circles show scores after training with written (upper, dark) or spoken (lower, light) captions. Panels show \textbf{(a)} noun neighbor purity, \textbf{(b)} synonym top-10 accuracy, and \textbf{(c)} agreement between word directions estimated from two sets of backgrounds (split-half cosine similarity). Scores use four word images, averaged over five randomly sampled typography settings. Gray vertical lines show scores before finetuning using all 96 word images. The dashed line marks chance noun purity.}
\label{fig:finetune}
\end{figure}

\paragraph{Objective.} A run trains one of the eleven self-supervised models end to end against a frozen encoder. The image side is the model's class token passed through a linear projection, without bias, to the encoder width of 1,024. The caption side is the frozen encoder's vector for the caption. A written caption is embedded by Qwen3-Embedding-0.6B \citep{zhang2025qwen3embedding}, read at the last token without an instruction prefix, and a spoken caption by w2v-BERT~2.0 \citep{seamlessm4t2023}, as the mean of its frames over the utterance. Both sides are scaled to unit length, and the loss is symmetric InfoNCE over a batch of 1,056 pairs, with a learned inverse temperature initialized at $1/0.07$ and capped at 100 \citep{radford2021clip}.

\paragraph{Optimization and schedule.} The optimizer is AdamW with $\beta = (0.9, 0.98)$, $\epsilon = 10^{-6}$, and weight decay 0.1, with the gradient norm clipped at 1.0. The learning rate is $5\times10^{-5}$ for the model and $5\times10^{-4}$ for the projection and temperature, with linear warmup and then cosine decay to zero, under bfloat16 mixed precision. Each training image is a random crop covering 90 to 100 percent of its area, resized to 224 pixels with bicubic interpolation and normalized with the ImageNet channel statistics, with no other augmentation. The schedule is three epochs of the filtered corpus, 1,310 steps (\tabref{tab:checkpoints}). Warmup takes a tenth of the schedule and the interval between validations on held-out pairs a twelfth, 131 and 109 steps. A run keeps the weights with the best validation loss and stops after three validations without improvement. The projection and temperature are discarded after training, so a finetuned model differs from its pretrained model only in its weights.

\paragraph{Corpus and text filter.} SPEECH-COCO \citep{havard2017speechcoco} pairs each COCO image \citep{lin2014coco} with about five synthesized spoken captions, each a reading of one of that image's COCO captions \citep{chen2015cococaptions} in one of eight named voices at one of three speaking rates. The speech run takes the recording and the written run the text of the same caption, so the two runs see the same images and captions. Every image is scanned with the CRAFT detector \citep{baek2019craft} at the settings used for the backgrounds, and any image with detected text is excluded, which leaves 461,195 captions, each with its recording. The filter is required because captions often transcribe writing in the image, and a model trained on such pairs could group rendered words by matching them to captions. Validation holds out two percent of the images with all their captions, so no image is on both sides.

\paragraph{Checkpoints.}
\begin{wraptable}{r}{0.50\textwidth}
\centering
\footnotesize
\setlength{\abovecaptionskip}{5pt}
\setlength{\belowcaptionskip}{\abovecaptionskip}
\setlength{\abovecaptionskip}{0pt}
\caption{The step of the checkpoint kept for each run. An asterisk marks an early stop.}
\label{tab:checkpoints}
\begin{tabular}{lrr}
\toprule
 & \multicolumn{2}{c}{SPEECH-COCO} \\
\cmidrule(l){2-3}
Model & written & speech \\
\midrule
\rowcolor{tabband}DINOv3-B & 1,199 & 1,308 \\
\rowcolor{tabband}DINOv3-L & 763$^{\ast}$ & 1,199 \\
\rowcolor{tabband}DINOv3-H & 436$^{\ast}$ & 981$^{\ast}$ \\
Web-DINO-1B & 218$^{\ast}$ & 545$^{\ast}$ \\
\rowcolor{tabband}DINOv2-B & 1,199 & 1,090 \\
\rowcolor{tabband}DINOv2-L & 981$^{\ast}$ & 981$^{\ast}$ \\
\rowcolor{tabband}DINOv2-G & 218$^{\ast}$ & 872$^{\ast}$ \\
MAE-B & 1,308 & 1,199 \\
MAE-L & 872$^{\ast}$ & 981$^{\ast}$ \\
MAE-H & 763$^{\ast}$ & 872$^{\ast}$ \\
\bottomrule
\end{tabular}
\end{wraptable}

\tabref{tab:checkpoints} gives the step of the checkpoint kept for every run. Every self-supervised model but DINOv3-7B is trained once against each of the two encoders, and every run keeps its weights from the step with the best validation loss. Four of the runs that stopped early did so at step 1,308, the last validation of the schedule.

\paragraph{Before and after finetuning.} \figref{fig:finetune} shows every finetuned model in three readouts before and after both runs. A finetuned model is measured exactly as its pretrained model, with the same word images, control images, and preprocessing. Word directions from four word images use the four backgrounds of one typography setting. The setting is drawn at random five times, with the same draws before and after finetuning, and the five draws give four distinct settings, one of them twice. A top-10 accuracy from four word images is the mean over the draws. In DINOv3-H the standard deviation of capital top-10 accuracy over the draws is 0.10 before finetuning, larger than its change after the written run. Mutual 10-NN alignment and linear CKA are computed against the same text and speech encoders as in \figref{fig:language_alignment}. 